\pdfoutput=1

\documentclass[10pt,twocolumn,letterpaper]{article}

\usepackage[pagenumbers]{cvpr}   %

\usepackage{multirow}
\usepackage{colortbl}
\usepackage{makecell}
\usepackage{pifont}
\usepackage{listings}
\ifdefined\pdfminorversion\pdfminorversion=7\fi

\definecolor{nvgreen}{HTML}{76B900}
\definecolor{nvdark}{HTML}{4A7A00}
\definecolor{oursrow}{HTML}{EDF5DC}
\definecolor{grayrow}{HTML}{F3F3F3}
\definecolor{graytext}{HTML}{7A7A7A}

\newcommand{\method}{Agentic LongLive\xspace}
\newcommand{\bench}{LongStory-Online\xspace}
\newcommand{\Gm}{\mathbf{G}}
\newcommand{\Mm}{\mathbf{M}}

\newcommand{\sig}{$^{\ast}$}
\newcommand{\best}[1]{\textbf{#1}}
\newcommand{\gr}[1]{{\color{graytext}#1}}

\newcommand{\down}{$\downarrow$}
\newcommand{\ours}{\rowcolor{oursrow}}

\definecolor{cvprblue}{rgb}{0.21,0.49,0.74}
\usepackage[pagebackref,breaklinks,colorlinks,allcolors=cvprblue]{hyperref}

\def\confName{CVPR}
\def\confYear{2027}

\providecommand{\linenumbers}{}
\providecommand{\nolinenumbers}{}

\newcommand*{\codeurl}{https://github.com/NVlabs/LongLive/tree/main/Agentic-LongLive}
\newcommand*{\projecturl}{https://nvlabs.github.io/LongLive/Agentic-LongLive/}

\title{Agentic LongLive: Asynchronous Agentic Memory \\ for Long Video Generation}

\hypersetup{
  pdftitle={Agentic LongLive: Asynchronous Agentic Memory for Long Video Generation},
  pdfauthor={Shaoteng Liu, Shuai Yang, Wei Huang, Luozhou Wang, Song Han, Yukang Chen}
}

\newcommand{\aff}[1]{\textsuperscript{#1}}
\DeclareMathSizes{8.7}{8.7}{6}{5}
\newcommand{\linkicon}[1]{\raisebox{-0.15em}{\includegraphics[height=1em]{arxiv/assets/#1}}\hspace{0.35em}}
\newcommand{\maybehref}[2]{\ifx#1\empty#2\else\href{#1}{\textcolor{nvgreen}{#2}}\fi}

\newcommand{\arxivauthors}{%
  \parbox{\linewidth}{\centering
    {\fontsize{9pt}{16pt}\selectfont\bfseries
      Shaoteng Liu\aff{3}, ~ Shuai Yang\aff{1}, ~ Wei Huang\aff{1}, ~ Luozhou Wang\aff{1}, ~ Song Han\aff{1\thinspace2}, ~ Yukang Chen\aff{1}\par}
    \vspace{2.2mm}
    {\fontsize{8.7pt}{11pt}\selectfont \aff{1}NVIDIA \quad \aff{2}MIT \quad \aff{3}CUHK\par}
    \vspace{10pt}
    {\fontsize{9pt}{11pt}\selectfont\bfseries\color{nvgreen}%
      \maybehref{\codeurl}{\linkicon{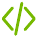}Code}%
      \hspace{1.8em}%
      \maybehref{\projecturl}{\linkicon{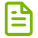}Project Page}\par}
  }%
}
\author{\arxivauthors}

\makeatletter
\def\@maketitle{%
  \newpage
  \null
  \vskip 0.1in
  \begin{center}
    {\color{nvgreen}\fontsize{17pt}{21pt}\selectfont\bfseries \@title\par}
    \vskip 11pt
    {\@author\par}
  \end{center}
  \vskip 14pt
}
\makeatother

\newcommand{\arxivdate}{\the\year-\the\month-\the\day}
\AddToShipoutPicture*{\AtTextUpperLeft{%
  \put(0,18){\includegraphics[width=90pt]{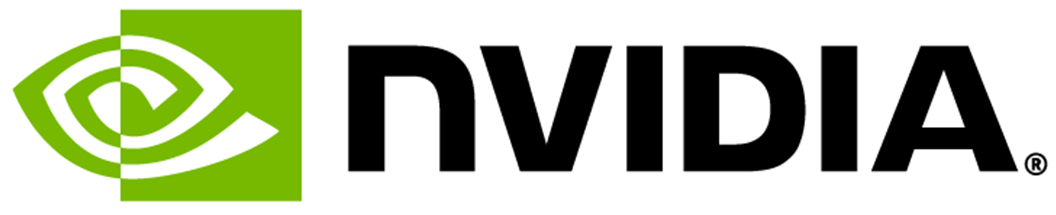}}%
  \put(\LenToUnit{\textwidth},18){\makebox[0pt][r]{\fontsize{8pt}{10pt}\selectfont\arxivdate}}%
  \put(0,10){\color{gray}\rule{\textwidth}{0.8pt}}%
}}

\begin{document}
\twocolumn[{%
\renewcommand\twocolumn[1][]{#1}%
\maketitle
\begin{center}
  \centering
  \captionsetup{type=figure}
  \vspace{-1.2em}
  \includegraphics[width=\textwidth]{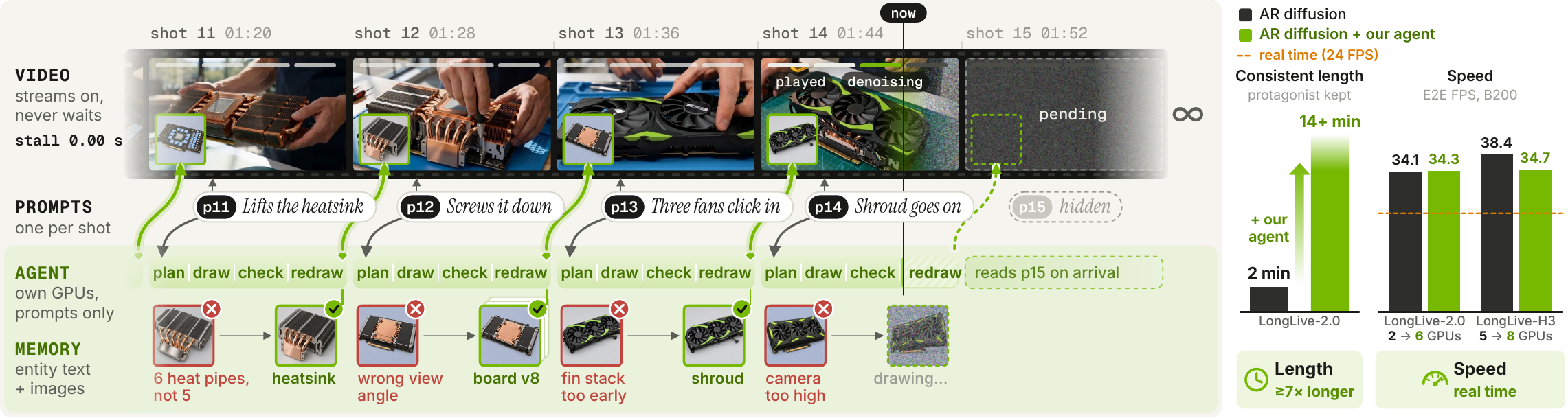}
  \vspace{-1.4em}
  \captionof{figure}{\textbf{\method keeps a live story consistent in real time.}
  \emph{Left:} a live 40-shot story (assembling a graphics card) during shot 14. Each prompt is revealed only when its shot starts, and the agent prepares memory one shot ahead and redraws failed images (red).
  \emph{Right:} with our agent, LongLive-2.0 keeps the protagonist recognizable for 14 minutes instead of 2 (\cref{fig:analysis}), and both LongLive-2.0 and LongLive-H3 still stream faster than real time on B200, with 3--4 more GPUs for the agent (\cref{tab:speed}).}
  \label{fig:teaser}
\end{center}
\vspace{0.6em}

}]
\begin{abstract}
Real-time autoregressive video generators stream minutes of video and accept new prompts on the fly, but struggle to keep people, objects, and places consistent over a long interactive story.
Agent systems provide such memory for offline multi-shot generation, yet they read the whole script in advance, plan each shot before generating it, and regenerate failed shots, which a live stream does not allow.
We present \method, an asynchronous agentic memory for long video generation.
A memory agent on separate GPUs reads only the prompts revealed so far, never the generated video, and keeps a text state and evolving images for every person, object, and scene.
From this entity memory it composes a global image and a per-shot memory image, which an auxiliary GPU prefills into the generator's attention sink before the next shot starts, and writes entity cards for the prompts that name them, so the video never waits for memory and video errors are not written back.
A small language model, trained with reinforcement learning and an image reward calibrated using downstream video outcomes, decides what to remember before the next prompt is known.
On \bench, a new benchmark of 25 stories (857 shots) revealed one shot at a time, \method improves persistent-state accuracy of LongLive-2.0, its base generator, by 26.3 points, protagonist identity by 44.9, and revisited-place consistency by 40.4, at about 1\% extra generation time on H100.
The full system streams faster than real time on one 8$\times$B200 node with the generator in BF16 (34.3\,FPS), keeps image quality stable over 14-minute stories, and improves identity on two MiniMax-H3 generators, one of which, LongLive-H3, also runs in real time with it.
To our knowledge, it is the first real-time video generator with text and image memory kept by an LLM agent.
\end{abstract}

\section{Introduction}
\label{sec:intro}

Video generation models are now strong enough to tell stories~\cite{wan2025wan,kong2024hunyuanvideo,polyak2024moviegen,minimax2026h3}, and agents that plan shots, track characters, and check results keep multi-shot videos consistent across scenes~\cite{wang2026storyengine,huang2026vimax,wu2025movieagent,zheng2024vgot,xie2024dreamfactory}.
Live streaming, games, and world models need a different kind of generator: autoregressive (AR), few-step models that stream long videos in real time and accept new prompts while they play~\cite{yin2025causvid,huang2025selfforcing,yang2025longlive,chen2026longlive2,liu2025rollingforcing,he2025matrixgame2,deepmind2025genie3,nvidia2026streamrender}.
These generators have no agent.
Their memory is the key/value (KV) cache: an attention sink that keeps the first frames~\cite{xiao2024streamingllm,yang2025longlive} and a short window of recent frames.
Older content is forgotten, and what is kept is the model's own output, so a wrong face or a lost object is copied forward~\cite{huang2025selfforcing,cui2025selfforcingpp}.

Adding an agent is hard because agents think slowly and real-time video cannot wait.
A live session imposes four constraints that offline agent systems do not face:
\textbf{(C1)} the future is unknown, so memory for a shot must be ready before its prompt arrives;
\textbf{(C2)} there is no length limit, so GPU memory must stay constant;
\textbf{(C3)} the deadline is hard, so the video never waits for memory construction: planning, drawing, and prefilling must stay hidden, and once a prompt arrives only a quick choice among prepared memory is affordable;
\textbf{(C4)} there is no rollback, so played shots cannot be repaired.
Existing video agents rely on what a live session lacks: they read the whole script (C1), plan each shot while generation waits (C3), and inspect and regenerate failed shots (C4)~\cite{wang2026storyengine,huang2026vimax,shi2025animaker}.

We present \textbf{\method}, an asynchronous agentic memory that gives real-time AR generators a long-term memory without slowing them down (\cref{fig:teaser}).
\textbf{Asynchronous memory:} a memory agent runs on its own GPUs and commits the memory for shot $i$ before prompt $i$ is revealed, and an auxiliary GPU prefills it into the generator's KV cache, so the agent's time is hidden behind generation.
\textbf{Open-loop entity memory:} the agent reads only revealed prompts, never the video, so video errors are not written back. It keeps a text state and evolving images for every person, object, and scene, composes them into a global image $\Gm$ and a shot memory $\Mm$ that enter the frozen generator as attention sinks, and appends text cards of the entities a prompt names.
\textbf{Learning what to remember:} before the next prompt is known, the agent must choose which entities to draw. We train a small language model for this with reinforcement learning (RL), using a hindsight reward whose image term is calibrated using downstream video outcomes.

We also build \bench, 25 stories (857 shots, about 150 minutes) with hidden annotations, where prompts are revealed one shot at a time and every shot is generated exactly once.
On LongLive-2.0~\cite{chen2026longlive2}, \method improves persistent state, protagonist identity, and location on scene changes by 26 to 45 points, adding about 1\% generation time on H100, and streams faster than real time on one 8$\times$B200 node in BF16 (34.3\,FPS).
Its image quality stays flat over 14-minute stories, its identity gains carry over to two few-step MiniMax-H3 generators~\cite{minimax2026h3,lu2026raven}, with LongLive-H3 also in real time on B200 (1.45$\times$), and RL lets an 8B planner replace the 35B planner.
We contribute:
\begin{itemize}[leftmargin=*,itemsep=0pt,topsep=2pt]
  \item an \textbf{asynchronous agentic memory system} that hides agent latency behind real-time generation with constant GPU memory and no length limit, to our knowledge the first real-time video generator with text and image memory kept by an LLM agent;
  \item an \textbf{open-loop two-layer memory} as attention sinks of a frozen AR generator, with entity cards in the prompt;
  \item an \textbf{RL recipe} with a video-calibrated reward that teaches a small LLM what to remember, and the \textbf{\bench} benchmark.
\end{itemize}
Related work is reviewed in App.~\ref{sec:related}.

\section{Method}
\label{sec:method}

\begin{figure}[t]
  \centering
  \includegraphics[width=\linewidth,trim=0 3.5 0 2.5,clip]{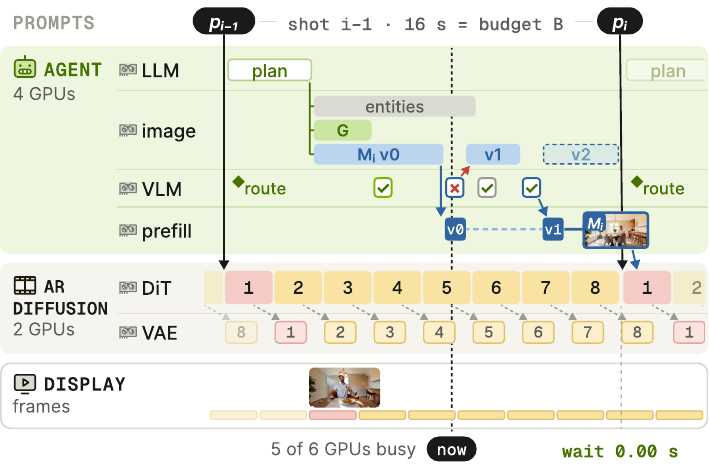}
  \caption{\textbf{System overview}: one shot of LongLive-2.0 on six H100s (agent steps to scale, decoding schematic).
  While shot $i{-}1$ is generated, the agent prepares the memory for shot $i$. Version 0 of $\Mm_i$ is prefilled at once, and a redraw replaces it only if it passes its check. At $p_i$, the DiT swaps in the newest prefilled version without waiting (v2, still being drawn, is dropped).}
  \label{fig:overview}
\end{figure}

\subsection{Online long-video generation}
\label{sec:setting}
A live session streams prompts $p_1, p_2, \dots$, one per shot.
As in streaming generation~\cite{yang2025longlive,chen2026longlive2}, $p_i$ is revealed only when shot $i$ starts.
The generator $\mathcal{F}$ produces shot $i$ from the prompt, its KV cache, and a memory $m_i$ from an agent $\mathcal{A}$:
\begin{equation}
  v_i = \mathcal{F}\big(p_i,\, \mathrm{KV}_i,\, m_i\big), \quad
  m_i = \mathcal{A}\big(p_{1:i-1},\, m_{1:i-1}\big).
  \label{eq:online}
\end{equation}
Memory for shot $i$ is committed before $p_i$ is revealed (C1) and never depends on generated video $v_{<i}$ (\emph{open loop}).
Each shot is generated once (C4).
The generator never waits for $\mathcal{A}$ to plan, draw, or prefill (C3). After $p_i$ arrives, it only selects which prefilled memory blocks to use (\cref{sec:memory}).
The attention window is fixed, so GPU memory stays constant for any length (C2).
If $m_i$ is not ready when shot $i$ starts, the generator reuses the newest committed memory $m_{i-k}$.
We call $k$ the \emph{memory lag}: how many shots old the memory is ($k{=}0$ when $m_i$ is ready in time).

\subsection{Asynchronous agentic memory}
\label{sec:async}

\paragraph{Ranks and one-way data flow.}
\emph{DiT ranks} generate video (\cref{fig:overview}).
\emph{Auxiliary ranks} run a decoder that streams latents through the VAE, like the asynchronous decoder of LongLive-2.0~\cite{chen2026longlive2}, and a prefill worker that holds a copy of the DiT weights, encodes memory images with the frozen VAE, and computes their sink K/V, each on its own GPU.
\emph{Agent ranks} host a vision--language model that plans and checks, and an image model that draws memory images.
Data flows one way: from the agent to the prefill worker, then to the DiT, and on to the decoder for display.
The agent never waits for decoding, and memory K/V are ready before the shot that uses them (GPUs per role in \cref{sec:setup}).

\paragraph{Commit and swap.}
Memory for shot $i$ is prepared while shot $i-1$ is generated, between the reveals of $p_{i-1}$ and $p_i$ (\cref{fig:overview}).
Version 0 is committed as soon as it is drawn, even if it fails its check, so checks do not cause lag ($\Gm$ follows a stricter rule, App.~\ref{sec:supp-speed}), and is prefilled at once.
If time remains, the agent may redraw an image that its check flags. The new version replaces the committed one only if it passes and scores no lower.
At the boundary, the DiT swaps in the newest fully prefilled version as a whole, without waiting for one in progress, so no shot mixes two versions. If no new version is ready, it keeps the previous memory (lag, \cref{sec:setting}).
The agent thus has one shot's time per memory (budget in App.~\ref{sec:supp-speed}), and as the video never waits, the generator alone sets the throughput.

\begin{figure*}[t]
  \centering
  \includegraphics[width=\linewidth,trim=0 4.5 0 8,clip]{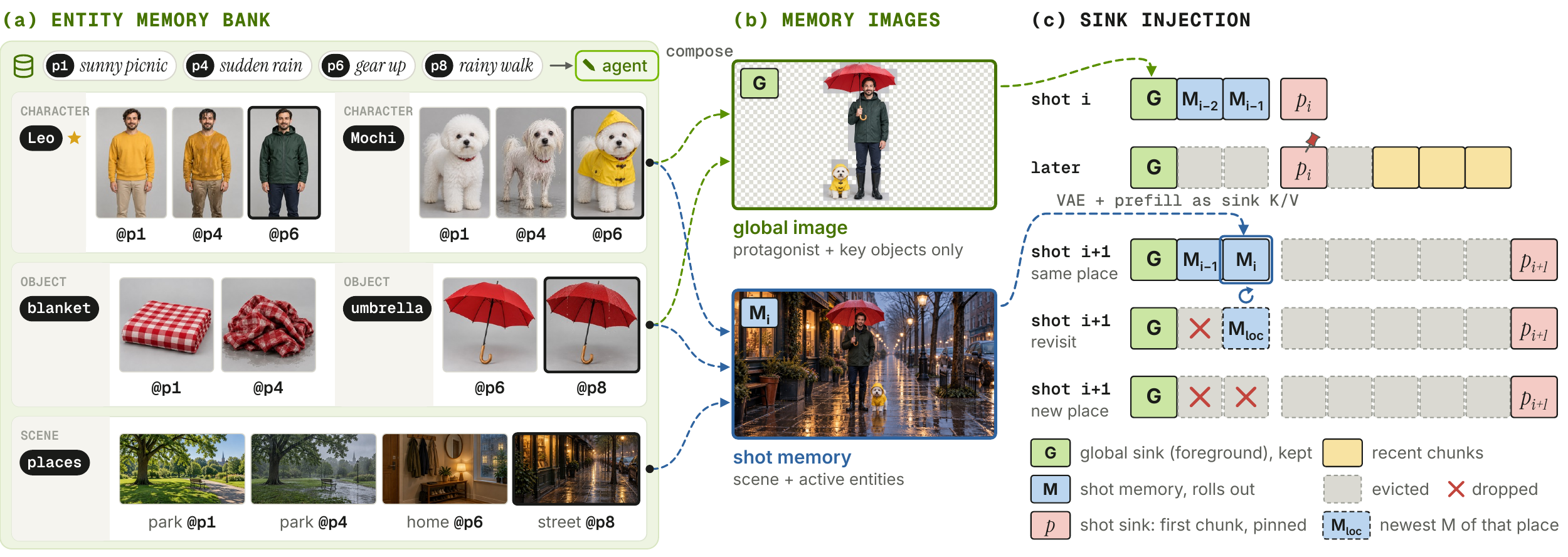}
  \caption{\textbf{Two-layer memory} on an illustrative story at $i{=}8$, where a sunny picnic turns to rain (images drawn with GPT Image).
  (a) Entity memory, each image tagged with the prompt that caused it (@$p$).
  (b) $\Gm$ and $\Mm_i$, composed from the newest images of the active entities (dark outline).
  (c) The DiT's KV window during shot $i$ and at the start of shot $i{+}1$ for each scene-routing choice.}
  \label{fig:memory}
\end{figure*}

\subsection{Two-layer memory}
\label{sec:memory}

\paragraph{Layer 1: entity memory.}
Every person, object, and scene is an entity with a text state and images (\cref{fig:memory}a): views for people and objects, and an empty plate for scenes.
After a prompt is revealed, the agent draws new or changed entities.
Since the next prompt is unknown (C1) and played shots are final (C4), memory holds the current belief about each entity.
Images keep only attributes that persist across shots (identity, clothing, dirt, damage, assembly) and leave pose and action to the next prompt.
Each state records its source, so a contradicting later prompt can revoke it (App.~\ref{sec:supp-memory}).

\paragraph{Layer 2: memory images.}
Generators cannot read an entity database, so the agent composes two images for them (\cref{fig:memory}b).
The global image $\Gm$ shows the protagonist and long-term key objects in their current state on a plain studio backdrop, and is refreshed when one of them changes.
The shot memory $\Mm_i$ shows the current scene and its active entities, one per shot.
$\Gm$ thus carries identity and state but no place, which is left to $\Mm$ and the prompt.
Both are composed with entity images as references rather than redrawn from text, so an object keeps its look across shots.
Entity images are drawn or edited in the background and checked once.
The rules for refining $\Gm$ and $\Mm$ are in App.~\ref{sec:supp-speed}.

\paragraph{Sink injection.}
$\Gm$ and $\Mm$ are encoded by the frozen VAE and prefilled as sink K/V (\cref{fig:memory}c).
Only the \emph{foreground} of $\Gm$ enters the sink (\cref{fig:memory}b): the agent segments its members and drops the backdrop, which MiniMax-H3 would otherwise copy into new scenes as a grey studio. \mbox{LongLive-2.0} keeps the whole $\Gm$ (App.~\ref{sec:supp-memory}).
$\Gm$ is conditioned on the description of its members and stays as the permanent global sink.
Each $\Mm$ is conditioned on the revealed prompt it was drawn from.
Shot $i$ starts from window $[\Gm, \Mm_{i-2}, \Mm_{i-1}]$ as a \emph{new shot}: an 8-frame gap follows the latest $\Mm$ and the first chunk is prefixed with ``The scene transitions.'', the multi-shot format of the base model~\cite{chen2026longlive2}.
That chunk is pinned as the shot sink, and the window rolls to $[\Gm$, first chunk, recent chunks$]$.
$\Mm$ thus sets the appearance at the start and then leaves, so the video follows the prompt instead of copying the layout of the memory image.
Each $\Mm$ is prefilled once while attending to $\Gm$ and earlier $\Mm$, so its K/V carry the memory history.
When $\Gm$ changes, the $\Mm$ blocks are prefilled again under the new $\Gm$.

\paragraph{Scene routing.}
$\Mm_{i-1}$ shows the previous location, but the next prompt may continue there, move to a new place, or return to an earlier one.
When $p_i$ is revealed, the agent's LLM reads it with the names and descriptions of the locations it has drawn and picks one of them or \emph{new}, with decoding constrained to these options.
If it picks the location of $\Mm_{i-1}$, the memory is kept.
If it picks another known location, the newest $\Mm$ drawn there is recalled and injected as $[\Gm, \Mm_{\text{loc}}]$.
For a new place, only $\Gm$ is injected (\cref{fig:memory}c).
The choice reads only revealed prompts and committed memory, and no memory is built after $p_i$ arrives.
The prefill worker keeps the newest $\Mm$ of every known location prefilled under the current $\Gm$, so a recall only swaps blocks.
On LongLive-2.0, routing is the only model call on the critical path (\cref{sec:speed}).

\paragraph{Entity cards.}
Images carry places and a look that words do not pin down, while the text encoder reads attribute words directly, so memory also enters as text.
When $p_i$ is revealed, every entity of the committed memory whose name occurs in $p_i$ (whole-word string match) gets a card appended after the prompt: its name, description, and persistent state (example in App.~\ref{sec:supp-memory}).
The agent writes these texts in advance, so only string matching remains after the prompt.
Like images, a card reflects memory up to shot $i{-}1$: a change stated in $p_i$ is in the prompt itself and enters the card from the next shot on.

\paragraph{Other generators.}
The same memory drives two chunked-AR versions of MiniMax-H3~\cite{minimax2026h3}, RAVEN~\cite{lu2026raven} and LongLive-H3 (\cref{sec:setup}), which take $\Gm$ and the routed $\Mm$ as sink chunks and entity cards in their prompt.
Where text enters decides the cost of memory. With cross-attention, as in LongLive-2.0, memory K/V depend on the memory images and their captions from earlier prompts, not on the newly revealed prompt, and are fully prefilled ahead of time. With text at the start of the sequence, as in MiniMax-H3, they depend on the prompt and are computed once after it arrives (\cref{sec:speed}).

\begin{figure}[t]
  \centering
  \includegraphics[width=\linewidth,trim=0 4.5 0 1,clip]{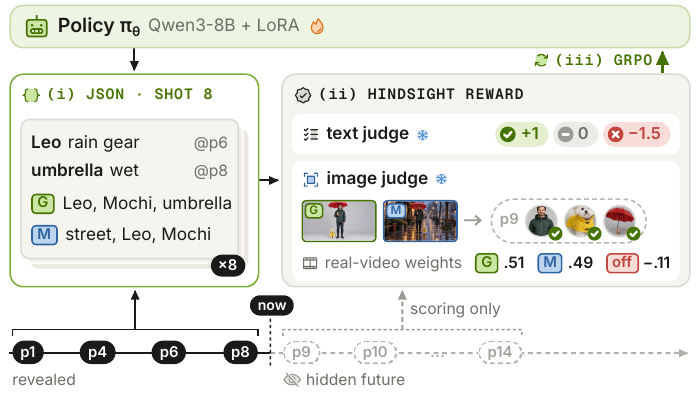}
  \caption{\textbf{Learning what to remember.} (i) A LoRA policy writes one JSON memory update per shot from the revealed prompts. (ii) The hidden future script is used only for scoring, by a text judge and an image judge with weights calibrated using downstream video outcomes. (iii) GRPO updates the policy.}
  \label{fig:rl}
\end{figure}

\subsection{Learning what to remember}
\label{sec:rl}
Memory images have few slots and the next prompt is unknown (C1): which entities should be drawn now, and which kept for later?
We train the planner with RL.
Only an LLM LoRA is trained, and the image, video, VAE, and judge models stay frozen (\cref{fig:rl}).

\paragraph{State and action.}
The policy sees the entity bank as text, the four newest prompts, and a slot budget.
For each shot it writes one JSON object with every text role of the agent: entity state updates, drawing prompts, the members of $\Gm$ ($\le$3) and $\Mm$ ($\le$4), and the sink captions (fields in App.~\ref{sec:supp-rl}).

\paragraph{Hindsight reward.}
After each decision, a scorer compares it with hidden annotations of later shots, which are used only for scoring.
The reward of a decision is
\begin{equation}
  r = 0.35\,r_{\text{text}} + 0.3\,r_{\text{img}} + 0.15\,r_{\text{prec}} + 0.05\,r_{\text{ret}} - r_{\text{pen}}.
  \label{eq:rimg}
\end{equation}
The \emph{text term} $r_{\text{text}}$ asks an LLM judge whether the persistent states that the next six shots need, but terse prompts will not restate, are remembered correctly, missing, or wrong. The terms $r_{\text{prec}}$ and $r_{\text{ret}}$ reward precise and retained facts, and $r_{\text{pen}}$ penalizes size, false facts, and length.
Recalled facts alone do not make better videos (\cref{sec:rl-exp}): the few image slots must also hold the entities that the next shot shows.
The \emph{image term} $r_{\text{img}}$ therefore renders $\Gm$ and $\Mm$ from the chosen members with the agent's image model, and a VLM checks whether each entity of the next shot is recognizable and correctly drawn.
It rewards each such entity in $\Gm$ or $\Mm$ and charges each member of $\Mm$ absent from the next shot.

\paragraph{Calibrating the image term with video outcomes.}
Generating videos in the RL loop is too slow, so we calibrate the image term on existing videos.
A regression of judged entity attributes in real videos on memory composition shows that an on-screen entity gains about as much state accuracy in $\Gm$ as in the latest $\Mm$, while each absent member of $\Mm$ costs a little. Normalizing these effects gives the weights of $r_{\text{img}}$ (App.~\ref{sec:supp-rl}).
A memory slot is thus valued by what it adds to the video, not by how many facts it recalls.

\paragraph{Training.}
A teacher plays 56 training stories online. Rejection-sampling fine-tuning~\cite{yuan2023rft} on its best outputs warm-starts the policy, and GRPO~\cite{shao2024deepseekmath} with each shot as a decision then optimizes this reward (App.~\ref{sec:supp-rl}).

\section{Experiments}
\label{sec:exp}

\subsection{Setup}
\label{sec:setup}

\paragraph{Protocol and benchmark.}
Every method sees one prompt per shot, generates each shot once, and never reads future prompts.
Regeneration and best-of-$N$ are not allowed, but a baseline keeps its own closed-loop mechanisms.
\bench has 25 stories, 857 shots, and about 150 minutes of video: nine stories written by hand (211 shots) and sixteen LongStory-Hard stories designed by an LLM with states accumulated by code (646 shots).
Stories have more entities than memory slots, long absences and revisits, persistent changes (clothing, dirt, damage, assembly), and prompts that overturn earlier inferences.
Each shot has a hidden annotation of entities, persistent states, and location, frozen before any method was run.
Prompts are \emph{terse}, like real user input: a place or person is described once, and later prompts use names or ``back in the kitchen''.

\paragraph{Baselines.}
\emph{LongLive-2.0}~\cite{chen2026longlive2} shares our backbone and keeps its built-in memory: sinks on the first chunk of the video and of each shot, and a KV window carried across shots.
Four open AR models on Wan2.1-T2V-1.3B, \emph{Self-Forcing}~\cite{huang2025selfforcing}, \emph{CausVid}~\cite{yin2025causvid}, \emph{Causal-Forcing}~\cite{zhu2026causalforcing}, and \emph{Rolling Forcing}~\cite{liu2025rollingforcing}, run one rollout per story with the prompt switched at each shot.
Self-Forcing breaks down within minutes, as in its official code (83\% of its shots have severe artifacts), so we do not rank it.
On MiniMax-H3~\cite{minimax2026h3}, we test two few-step chunked-AR generators: \emph{RAVEN}~\cite{lu2026raven} and \emph{LongLive-H3}.
We distill LongLive-H3 from MiniMax-H3 by causal distribution matching distillation (DMD)~\cite{yin2024dmd} on prompts alone. It generates video and audio in chunks of about 0.7\,s, each attending to the text, the first chunk, and the two previous chunks (App.~\ref{sec:supp-memory}).
Without memory, each 8\,s shot is generated from its prompt alone. A second baseline carries K/V across shots as LongLive-2.0 does (App.~\ref{sec:supp-memory}).
The non-real-time systems \emph{StoryMem}~\cite{zhang2025storymem} and \emph{StoryEngine}~\cite{wang2026storyengine}, an offline variant of ours, SkyReels-V2~\cite{chen2025skyreelsv2}, and all baseline adaptations are in App.~\ref{sec:supp-baselines}.

\paragraph{Metrics.}
A VLM judge (Gemini 3.8 Flash), different from every model of the real-time agent and of RL training, compares four frames of each shot with its annotation: \emph{State} (persistent attributes correct), \emph{Identity} (the visible protagonist matches its description), and \emph{Location} (on scene changes).
\emph{Same protagonist} compares each shot whose plan shows the protagonist with the method's first shot of it (0 if the protagonist is missing).
\emph{Revisit} rates whether a return shows the same place as the last visit and counts only returns where both shots are at their planned place~\cite{wang2026storyengine}.
We also report VBench imaging quality~\cite{huang2024vbench} and the share of shots with severe artifacts.
Framing, imaging drift~\cite{liu2025rollingforcing}, and ArcFace face identity (MIR~\cite{wang2026storyengine}) are in App.~\ref{sec:supp-bench} and~\ref{sec:supp-limits}.
Re-judging agrees on 92--96\% of decisions. As the same model also wrote the Hard stories, a second judge from another family (GPT-5.5) re-judged 14 runs and confirms 87 of the 94 significant comparisons, with no sign flip (App.~\ref{sec:supp-bench}).
Automated-metric differences quoted in the text are differences of table entries, significant unless noted (a story-clustered bootstrap 95\% interval excludes zero).

\paragraph{Implementation.}
LongLive-2.0 runs Wan2.2-TI2V-5B~\cite{wan2025wan} in BF16 with 4 steps and 10.5\,s shots.
The agent checks images and routes scenes with Qwen3.6-35B-A3B~\cite{qwen2026qwen36} and draws with FLUX.2-klein-4B~\cite{bfl2026flux2klein}, each on one H100.
In the full system, the RL planner (Qwen3-8B~\cite{yang2025qwen3} with LoRA~\cite{hu2022lora}) writes the memory on a third H100. Without RL, the 35B model also plans.
The full system thus uses six H100s (DiT 1, auxiliary 2, agent 3), against two for LongLive-2.0 alone (DiT and asynchronous decoder).
Memory streams come from the live agent against a stand-in generator with the same shot time (zero lag on 97--99\% of shots) and are replayed into each generator, pixel-identical to a live run.
On LongLive-2.0, rows with the RL planner pair each of three planners, trained with different seeds, with one video seed.

\begin{table*}[t]
\centering
\caption{\textbf{Main results on \bench} (online protocol, terse prompts). \emph{Imaging}: VBench imaging quality. \emph{Artifacts}: share of shots with severe artifacts. \emph{Time}$^\dagger$: seconds per shot (10--10.5\,s of video on Wan, 8\,s on H3) with the released code \emph{before speed optimization}, in BF16 on H100 (four GPUs per video on H3), including decoding. \emph{Full system}: RL planner, LLM scene routing, and entity cards.}
\label{tab:main}
\setlength{\tabcolsep}{4.6pt}
\renewcommand{\arraystretch}{0.92}
\resizebox{\linewidth}{!}{%
\begin{tabular}{l ccccc cc l}
\toprule
Method & State & Identity & Same prot. & Location & Revisit & Imaging & Artifacts\down & Time$^\dagger$ / shot  \\
\midrule
\midrule
\multicolumn{9}{l}{\textit{Real-time on Wan} \quad (25 stories, 857 shots, 3 seeds)} \\
Causal-Forcing~\cite{zhu2026causalforcing} & 27.6 & 59.8 & 20.8 & 60.4 & 11.7 & 0.715 & 7.1 & 17.4  \\
Self-Forcing~\cite{huang2025selfforcing} & 33.1 & 50.3 & 12.0 & 10.6 & 4.0 & 0.594 & 83.4 & 18.2  \\
CausVid~\cite{yin2025causvid} & 23.3 & 41.2 & 17.5 & 33.1 & 7.4 & 0.713 & 19.0 & 13.4  \\
Rolling Forcing~\cite{liu2025rollingforcing} & 30.4 & 58.3 & 25.3 & 33.1 & 17.7 & \best{0.748} & 13.5 & 19.3  \\
\arrayrulecolor{black!25}\specialrule{0.3pt}{0.3ex}{0.3ex}\arrayrulecolor{black}
LongLive-2.0~\cite{chen2026longlive2} (built-in memory) & 38.2 & 44.1 & 18.1 & 68.8 & 15.0 & 0.667 & 7.8 & 15.8  \\
\ours \quad Ours: full system & \best{64.5} & \best{89.0} & \best{46.8} & \best{98.0} & \best{55.4} & 0.729 & \best{6.0} & 16.0 (+0.9\%)  \\
\midrule
\midrule
\multicolumn{9}{l}{\textit{Few-step AR on MiniMax-H3} \quad (25 stories, 857 shots, 2 seeds; LongLive-H3 real time after optimization, \cref{tab:speed})} \\
RAVEN~\cite{lu2026raven} (independent shots) & 36.1 & 61.0 & 24.5 & 84.9 & 26.5 & 0.700 & \best{0.5} & 44.1  \\
\quad + K/V carried across shots & 38.7 & 91.2 & 47.8 & 25.9 & 15.9 & 0.698 & 9.4 & 44.4 (+0.6\%)  \\
\ours \quad Ours: full system & \best{63.3} & \best{98.0} & \best{70.1} & \best{89.1} & \best{47.4} & \best{0.753} & 3.0 & 45.3 (+2.6\%)  \\
\arrayrulecolor{black!25}\specialrule{0.3pt}{0.3ex}{0.3ex}\arrayrulecolor{black}
LongLive-H3 (independent shots) & 35.0 & 58.2 & 20.9 & 85.6 & 24.8 & 0.676 & \best{0.9} & 28.1  \\
\quad + K/V carried across shots & 29.7 & 76.0 & 29.3 & 20.3 & 11.0 & 0.676 & 11.6 & 29.0 (+3.5\%)  \\
\ours \quad Ours: full system & \best{61.1} & \best{96.8} & \best{62.6} & \best{87.5} & \best{42.9} & \best{0.740} & 5.0 & 29.1 (+3.9\%)  \\
\bottomrule
\end{tabular}}
\end{table*}

\subsection{Main results}
\label{sec:main}

\begin{figure}[t]
  \centering
  \includegraphics[width=\linewidth,trim=0 2 0 2.5,clip]{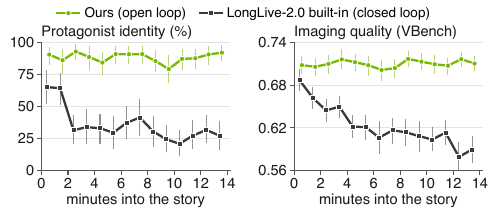}
  \caption{\textbf{Long horizon} on nine hand-written stories extended to 80 shots (14 min), two seeds. Per-minute points, 95\% intervals.}
  \label{fig:analysis}
\end{figure}

\paragraph{Same backbone.}
On LongLive-2.0, our system and the built-in memory share the generator and the prompts, so the comparison isolates memory.
The full system improves all five consistency columns (\cref{tab:main}). State accuracy rises by 26.3 points, and revisit consistency from 15.0\% to 55.4\%.
Imaging quality also improves and, unlike with the built-in memory, does not decay over a story (\cref{fig:analysis}).
The cost is about 1\% generation time on H100 (\cref{sec:speed}).
\Cref{sec:ablation,sec:rl-exp} separate the contributions of memory images, scene routing, entity cards, and the RL planner.

\paragraph{Other real-time AR models.}
The four Wan2.1 models run a smaller backbone at a lower resolution, so this comparison mixes backbone and memory.
Without our memory, LongLive-2.0 already leads them in state and location, but not in identity, same protagonist, or revisit (\eg 44.1\% \vs 59.8\% identity for Causal-Forcing).
With our memory it leads in all five consistency columns (\cref{tab:main}, top, all significant).
Rolling Forcing has the highest imaging quality (0.748) but rarely follows a scene change (33.1\% location).

\paragraph{Other backbones.}
On the MiniMax-H3 generators, our memory mainly improves identity: it rises from 61.0\% to 98.0\% on RAVEN and from 58.2\% to 96.8\% on LongLive-H3. State, same protagonist, and revisit rise with it, and location stays at or above the level without memory.
Without memory, severe artifacts are rare (0.5\% and 0.9\%) because each shot starts from noise, so errors cannot carry over.
Carrying K/V across shots, as LongLive-2.0 does, also raises identity (RAVEN $61.0\% \rightarrow 91.2\%$), but each shot then continues the previous one: location falls to 20--26\%, errors accumulate along the story (9.4\% and 11.6\% severe artifacts), and our full system leads it by 63--67 location points.
Memory or entity images given as references instead of sinks cause more severe artifacts, and entity references need a prompt rewrite after each prompt (App.~\ref{sec:supp-memory}).

\paragraph{Non-real-time references.}
On the same Wan2.2-5B backbone and keyframe model, our offline variant (the same memory with API models and synchronous refinement) gains 9.3 state points over StoryEngine but trails it by 15.7 revisit points (App.~\ref{sec:supp-baselines}).
On LongLive-2.0, our real-time full system reaches 93\% of StoryEngine's state accuracy and identity at 1/47 of its time per second of video.

\paragraph{User study.}
Thirty-two participants each evaluated ten stories, viewing four shots spread over each story, and picked the best version on four criteria (App.~\ref{sec:supp-userstudy}).
For persistent state, our full system receives a story-averaged best-version vote share of 98.8\% against the built-in memory of LongLive-2.0, and 97.8\% in a three-way choice with the independent-shot and carried-K/V versions of LongLive-H3 (ties split the vote). Identity and location receive 94 to 100\%.
Ours also receives every visual-quality vote on LongLive-2.0, whereas on LongLive-H3 these votes split between independent shots (55.1\%) and ours (43.6\%).

\begin{figure*}[t]
  \centering
  \includegraphics[width=\linewidth]{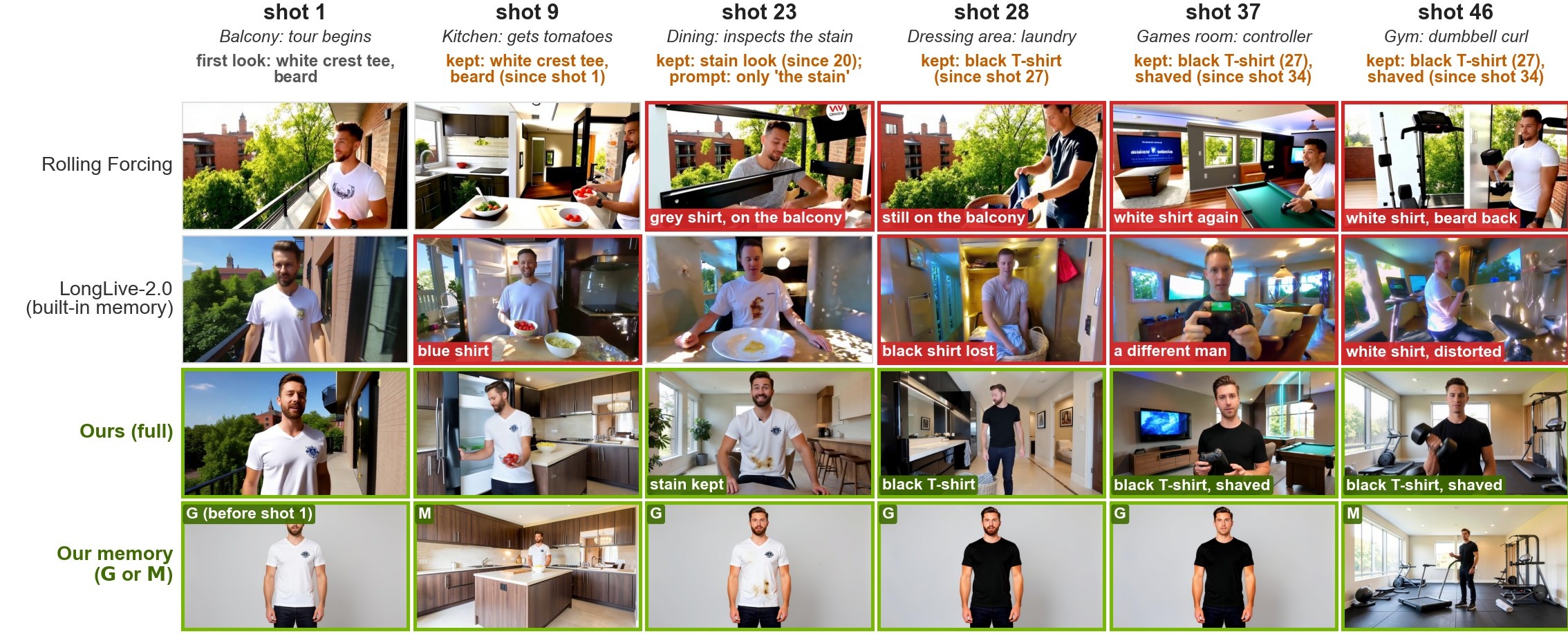}
  \caption{\textbf{Qualitative comparison} of real-time Wan-based models on a 48-shot home tour outside \bench (same protocol, terse prompts, 10.5\,s shots, seed 0). The presenter stains his shirt, changes into a black T-shirt, and shaves, and later prompts do not restate these changes (\emph{kept}). Each column is one shot, sampled at the same time in every row. Red tags mark baseline errors. Bottom row: our memory, which shows each state before the shot that needs it. More cases are on the \href{\projecturl}{project page}.}
  \label{fig:qual}
\end{figure*}

\paragraph{Qualitative results and long horizon.}
In \cref{fig:qual}, only our memory carries the presenter's stain, black T-shirt, and shave through the eight-minute tour.
Over 14 minutes (\cref{fig:analysis}, state in App.~\ref{sec:supp-memory}), our identity and imaging quality stay flat, whereas LongLive-2.0 loses identity after two minutes and imaging quality throughout.

\subsection{Speed}
\label{sec:speed}

\begin{table}[t]
\centering
\caption{\textbf{Speed.} \emph{FPS}: end to end, with decoding and memory (playback: 24\,FPS). \emph{Wait}: mean (max) stall per shot boundary. \emph{Overhead}: time added by memory. \emph{GPUs}: DiT / auxiliary / agent. \emph{Faster kernels} keep quality (blind pairwise judge, App.~\ref{sec:supp-speed}). Other configurations, quality, and lag: App.~\ref{sec:supp-speed}.}
\label{tab:speed}
\setlength{\tabcolsep}{2.5pt}
\footnotesize
\begin{tabular}{l cccc}
\toprule
Method & FPS & Wait (s) & Overhead & GPUs \\
\midrule
\multicolumn{5}{l}{\textit{LongLive-2.0, H100, release inference (BF16)}} \\
LongLive-2.0 & 16.0 & 0.00 & -- & 1/1/0 \\
Synchronous agent & 10.0 & 9.40 (16.6) & +60.4\% & 1/2/3 \\
\ours Ours (\cref{tab:main}) & 15.9 & 0.00 & +0.9\% & 1/2/3 \\
\midrule
\multicolumn{5}{l}{\textit{LongLive-2.0, 8$\times$B200, BF16}} \\
Release inference & 21.7 & 0.00 & -- & 1/1/0 \\
Faster kernels & 34.1 & 0.00 & -- & 1/1/0 \\
\ours Ours, faster kernels & 34.3 & 0.00 & $-$0.7\% & 1/2/3 \\
\midrule
\multicolumn{5}{l}{\textit{LongLive-H3, 8$\times$B200}} \\
Release inference & 10.0 & 0.00 & -- & 4/0/0 \\
Faster kernels & 38.4 & 0.00 & -- & 4/1/0 \\
\ours Ours, faster kernels & 34.7 & 0.00 & +10.4\% & 4/1/3 \\
\bottomrule
\end{tabular}
\end{table}

\Cref{tab:speed} starts from the release inference of \cref{tab:main}, adds our agent, then the speed optimizations that keep quality.

\paragraph{Does the video wait for the agent?}
A \emph{synchronous agent}, which builds memory only after a shot's prompt arrives, holds the video for 9.4\,s per shot (+60\%).
Our agent builds memory ahead of the prompt on its own GPUs and costs 0.9\% with no stall (waiting for refined memory or prefilling on the DiT GPU would cost 2.3\% and 6.9\%, App.~\ref{sec:supp-speed}).

\paragraph{Real time on LongLive-2.0.}
One H100 stays below real time even when optimized (0.91$\times$ without memory).
On B200, cuDNN attention and decoding overlapped across shots bring the full system to 34.3\,FPS (1.43$\times$) in BF16 on six GPUs with no measurable overhead, and every memory update is prefilled before the next shot (App.~\ref{sec:supp-speed}).

\paragraph{Real time on LongLive-H3.}
MiniMax-H3 is a partial exception to C3: its text comes first in the sequence, so encoding the chosen pictures and one forward pass stay on the critical path (1.0--1.2\,s per shot with the release code, 0.5\,s after optimization), while planning and drawing stay hidden.
Half of its release time is decoding after the whole shot. Faster DiT kernels and decoding split over the same four B200s bring a shot to 6.4\,s (30\,FPS), and streaming the decoder on a fifth GPU to 5.5\,s (34.7\,FPS, 1.45$\times$), with the first frames after 1.5\,s.
The agent on the other three GPUs has every update ready at the 8\,s playback pace.
On H100, generation needs eight GPUs (7.1\,s per shot), and an agent on H100 misses two thirds of the deadlines (App.~\ref{sec:supp-speed}).

\subsection{Ablation: memory design}
\label{sec:ablation}

\begin{table}[t]
\centering
\caption{\textbf{Memory design ablation} on LongLive-2.0 (25 stories, 3 seeds). The reference is the second row: $\Gm$ and the two newest $\Mm$ with LLM scene routing, written by the 35B planner. Each group changes one part of it. Paired differences: App.~\ref{sec:supp-memory}.}
\label{tab:ablation}
\setlength{\tabcolsep}{4pt}
\footnotesize
\begin{tabular}{@{}l cccc@{}}
\toprule
Design & State & Ident. & Loc. & Rev. \\
\midrule
LongLive-2.0, built-in memory & 38.2 & 44.1 & 68.8 & 15.0 \\
Image memory, 35B planner & 47.0 & 62.7 & 96.4 & 41.1 \\
\midrule
\multicolumn{5}{@{}l}{\textit{Fewer memory images}} \\
Static identity image only & 38.8 & 49.8 & 86.8 & 19.5 \\
$\Gm$ only & 40.1 & 46.1 & 84.2 & 18.4 \\
$\Gm$ + newest $\Mm$ & 45.6 & 57.9 & 96.0 & 41.0 \\
\midrule
\multicolumn{5}{@{}l}{\textit{Scene changes without LLM routing}} \\
Always keep the last $\Mm$ & 47.5 & 65.1 & 65.9 & 12.6 \\
Keyword gate (keep or drop) & 45.7 & 60.0 & 78.7 & 17.7 \\
\midrule
\multicolumn{5}{@{}l}{\textit{Adding cards of all injected members}} \\
Cards only, no images & 59.1 & 86.5 & 64.8 & 19.9 \\
Image memory + cards & 63.4 & 85.3 & 96.0 & 45.1 \\
\midrule
\multicolumn{5}{@{}l}{\textit{RL planner instead of 35B}} \\
Image memory, RL planner & 50.6 & 69.7 & 97.1 & 45.4 \\
\ours + named-entity cards (full system) & \best{64.5} & \best{89.0} & \best{98.0} & \best{55.4} \\
\bottomrule
\end{tabular}
\end{table}

\Cref{tab:ablation} studies memory design on LongLive-2.0: the rows with the 35B planner replay one fixed memory stream and change only which memory enters the generator and how, and the last group uses the streams of the three RL planners.
Image memory written by the 35B planner adds 8.8 state and 18.6 identity points to the built-in memory.

\paragraph{Shot memory.}
A static identity image or an updated $\Gm$ alone recovers little of this gain.
Most of it comes from $\Mm$, which shows the current scene and its entities: adding the newest $\Mm$ to $\Gm$ raises state to 45.6 and identity to 57.9, and revisit from 18.4\% to 41.0\%.
$\Gm$ matters through what the planner puts in it: an entity in the RL planner's $\Gm$ gains 7.9 points of state accuracy (App.~\ref{sec:supp-rl}).

\paragraph{Scene changes.}
Scene routing matters mainly for places: always keeping the last $\Mm$ lowers location from 96.4\% to 65.9\%, and a keyword gate, which can drop $\Mm$ at a new place but cannot recall an earlier visit, also falls short.

\paragraph{Images and text.}
Cards for all members of $\Gm$ and $\Mm$, used without images, recover persistent attributes but leave location and revisit near the built-in memory. Adding image memory restores these spatial gains, so the two forms of memory are complementary.
Cards for all members are long and dilute place words, so the full system writes cards only for the entities that the prompt names. On the RL planner's stream, they add 13.9 state and 19.3 identity points to image memory while location stays level.
They are not just more text: with the same image memory, they beat a naive history of the earlier sentences naming each entity by 5.0 state points (seed 0, App.~\ref{sec:supp-memory}).

\subsection{Ablation: learning what to remember}
\label{sec:rl-exp}

\begin{table}[t]
\centering
\caption{\textbf{Memory planners} on \bench. \emph{Text layer}: 857 decisions judged against hidden annotations. \emph{Recall}: needed facts stated correctly. \emph{In $\Mm$}: next-shot entities drawn in $\Mm$. \emph{Retain}: returning entities still in the entity bank. \emph{Video}: LongLive-2.0 with image memory, difference to the deployed 35B planner on the same stories and seeds. \gr{Gray}: not significant. RL rows average 3 training seeds (text reward only: 2). More columns: App.~\ref{sec:supp-rl}.}
\label{tab:rl}
\setlength{\tabcolsep}{3pt}
\footnotesize
\begin{tabular}{@{}l ccc cc@{}}
\toprule
& \multicolumn{3}{c}{Text layer (\%)} & \multicolumn{2}{c}{Video $\Delta$} \\
\cmidrule(lr){2-4}\cmidrule(l){5-6}
Planner & Recall & In $\Mm$ & Retain & State & Ident. \\
\midrule
Qwen3-8B, frozen & 57.7 & 35.4 & 60.5 & \gr{$-$1.6} & $-$4.9 \\
\quad + SFT (warm start) & 61.1 & 39.3 & 28.0 & \gr{$-$1.1} & \gr{$-$0.6} \\
Qwen3.6-35B (deployed) & 62.4 & 40.0 & 22.2 & -- & -- \\
RL, text reward only & \best{64.3} & 19.2 & \best{92.1} & $-$4.8 & $-$14.1 \\
RL, calibrated, not rendered & 60.2 & 66.6 & 81.2 & $+$2.5 & $+$5.0 \\
\ours RL, video-calibrated & 61.8 & \best{67.4} & 80.1 & \best{$+$3.6} & \best{$+$7.0} \\
\bottomrule
\end{tabular}
\end{table}

We train Qwen3-8B on 56 generated stories and test it on \bench, which other authors and models wrote (\cref{tab:rl}).
Remembering more in text does not give better videos.
With a text-only reward, the planner recalls the most facts and keeps the most returning entities, yet its videos have the lowest identity of all planners, 14.1 points below the 35B planner.
Its $\Mm$ contains only 19.2\% of the entities that the next shot shows.
With the video-calibrated image term of \cref{eq:rimg}, this share rises to 67.4\%, and image memory gains 3.6 state and 7.0 identity points over the 35B planner.
SFT on the same stories, the warm start of RL, brings the 8B planner level with the 35B planner, so the gain comes from RL.
Using the calibrated weights without rendering $\Gm$ and $\Mm$ during training keeps most of it ($+2.5$ state, $+5.0$ identity).
RL also keeps far more returning entities in the entity bank (83.9\% \vs 7.1\% after 8+ absent shots, \cref{fig:supp-retention}) and draws them once they return, whereas the 35B planner retires entities that leave the story.
Entity cards already state the tracked facts in text, so on top of them the RL planner's lead shrinks to $+0.5$ state, $+1.8$ identity, and $+3.2$ revisit points, none significant (App.~\ref{sec:supp-memory}). In the full system, RL thus lets one 8B model select members and write entity states in place of the 35B planner.

\subsection{Other applications}
\label{sec:other}

\begin{figure}[t]
  \centering
  \setlength{\tabcolsep}{0pt}
  \setlength{\fboxsep}{0.8pt}
  \newcommand{\rh}{0.127\linewidth}
  \definecolor{oursgreen}{RGB}{70,101,21}
  \newcommand{\flab}{\sffamily\fontsize{6}{7}\selectfont}
  \newcommand{\stamp}[2]{\makebox[0pt][l]{\hspace{#1}\raisebox{1.2pt}{\hspace{1.2pt}\colorbox{white}{\sffamily\fontsize{5}{6}\selectfont #2}}}}
  \newcommand{\obs}[2]{\makebox[\dimexpr\rh*480/264\relax][l]{\rlap{\includegraphics[height=\rh]{#1}}\stamp{0pt}{#2}}}
  \newcommand{\gen}[1]{\makebox[\dimexpr\rh*1456/264\relax][l]{\rlap{\includegraphics[height=\rh]{#1}}\stamp{0pt}{14.9\,s}\stamp{\dimexpr\rh*488/264\relax}{16.7\,s}\stamp{\dimexpr\rh*976/264\relax}{18.5\,s}}}
  \begin{tabular}{@{}l@{\hspace{2pt}}!{\color{black!40}\vrule}@{\hspace{2pt}}c@{\hspace{1.5pt}}c@{}}
    {\flab\itshape Shared input} & & {\flab\itshape Generated from 13.2\,s, camera moves back} \\[1pt]
    \obs{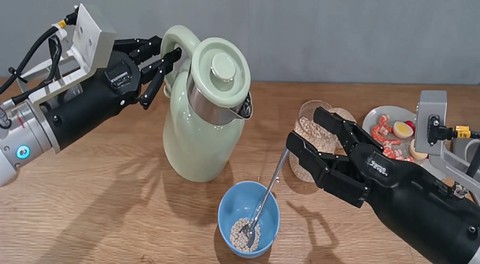}{0\,s} &
    \rotatebox{90}{\makebox[\rh][c]{\flab\color{black!80}Original}} & \gen{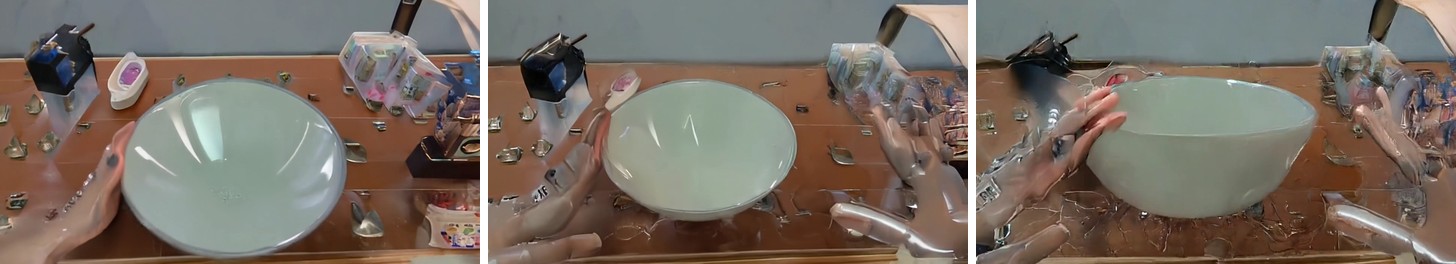} \\[1pt]
    \obs{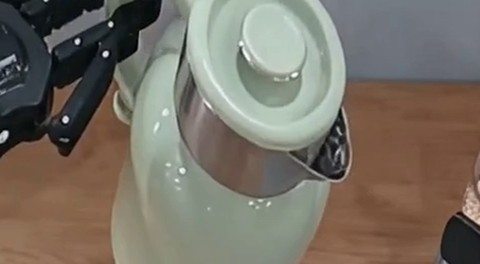}{2.5--13.2\,s} &
    \rotatebox{90}{\makebox[\rh][c]{\flab\bfseries\color{oursgreen}Ours}} & \gen{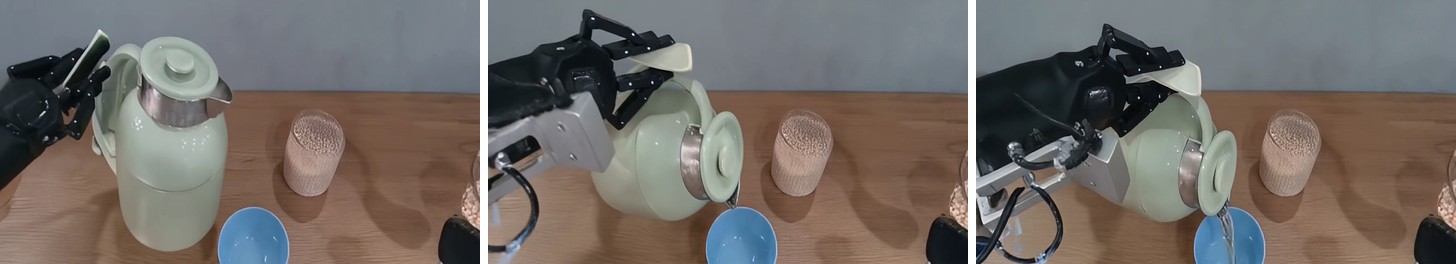} \\
  \end{tabular}
  \caption{\textbf{Robot video model}~\cite{huang2026longwam}. Both rows see the same input (left), whose close-up pushes the scene out of the context window.}
  \label{fig:robot}
\end{figure}

Because the agent only decides what the generator sees and leaves the generator unchanged, we also test it on robot video models (App.~\ref{sec:supp-plug}).
In robotics, a camera that stays close to an object for longer than the context window loses the rest of the scene.
On the robot video model of Long-WAM~\cite{huang2026longwam}, the original model then turns to a different scene. An observation-based variant of our memory, which stores the initial scene and an object crop in the sink, keeps both and brings them back (\cref{fig:robot}).
This variant differs from the agent of \cref{sec:method}: its memory pictures are frames that the camera has already shown, not drawn images. For long robot tasks (App.~\ref{sec:supp-plug}), it also reads the generated video (closed loop) and sees each instruction one segment before it plays, because a robot acts in one continuous scene whose state depends on what was generated.

\section{Conclusion}
\label{sec:conclusion}
We presented \method, an asynchronous agentic memory for real-time long video generation.
A memory agent that reads only revealed prompts keeps entity memory and commits memory images that an auxiliary GPU prefills into the generator's attention sink, so the video never waits for memory, and RL teaches a small planner what to remember.
On \bench, this improves state, identity, and place consistency of LongLive-2.0 at about 1\% extra generation time and streams faster than real time in BF16 on one B200 node. Image quality stays flat over 14-minute stories, and the identity gains carry over to two MiniMax-H3 generators, with LongLive-H3 also in real time.

\paragraph{Limitations.}
Memory needs extra GPUs for the agent ranks and the prefill worker (\cref{tab:speed}). On LongLive-H3, it also leaves per-shot visual quality to the base model: human raters judge it no better than that of independent shots.

{
    \small
    \bibliographystyle{ieeenat_fullname}
    \bibliography{main}
}

\clearpage
\twocolumn[{\centering\Large\textbf{\thetitle}\\\vspace{0.5em}Appendix\\\vspace{1.0em}}]
\appendix
\renewcommand{\thesection}{\Alph{section}}
\counterwithin{table}{section}
\counterwithin{figure}{section}
\renewcommand{\thetable}{\thesection\arabic{table}}
\renewcommand{\thefigure}{\thesection\arabic{figure}}

\noindent The appendix first reviews related work (\cref{sec:related}) and then follows the order of the experiments: system and speed (\cref{sec:supp-speed}, for \cref{sec:speed}), memory design (\cref{sec:supp-memory}, for \cref{sec:ablation}), memory planner (\cref{sec:supp-rl}, for \cref{sec:rl-exp}), benchmark and metrics (\cref{sec:supp-bench}), and baselines (\cref{sec:supp-baselines}), followed by tests of the agent on robot video models (\cref{sec:supp-plug}), by entity evolution, failure cases, and limitations (\cref{sec:supp-limits}), and by the human preference study (\cref{sec:supp-userstudy}).
Several analyses in this appendix evaluate the image memory without entity cards. There, \emph{image system} denotes the image memory with LLM scene routing and the RL planner (the RL planner's image-memory row of \cref{tab:ablation}), and \emph{full system} adds the cards.

\section{Related Work}
\label{sec:related}

\paragraph{Real-time AR video generation.}
Diffusion Forcing~\cite{chen2024diffusionforcing} and CausVid~\cite{yin2025causvid} make video diffusion causal.
Self Forcing~\cite{huang2025selfforcing} and its successors~\cite{cui2025selfforcingpp,liu2025rollingforcing,lu2026raven} close the train--test gap with self rollouts and few-step distillation.
MAGI-1~\cite{magi2025}, SkyReels-V2~\cite{chen2025skyreelsv2}, Krea Realtime~\cite{krea2025realtime}, APT2~\cite{lin2025apt2}, and StreamDiT~\cite{kodaira2026streamdit} scale this up.
To run without a length limit, these models keep a fixed window of recent frames and an attention sink~\cite{xiao2024streamingllm}.
Rolling Sink~\cite{li2026rollingsink} keeps this cache close to what the model saw in training, without retraining, so rollouts stay stable far beyond the training horizon.
LongLive~\cite{yang2025longlive} adds a frame sink and KV re-caching for prompt switches.
LongLive-2.0~\cite{chen2026longlive2} adds multi-shot sinks and parallel inference, and LongLive-Plug~\cite{yang2026longliveplug} distills few-step and long-video LoRAs once and plugs them into other models of the same backbone family.
Other works pack or route longer context~\cite{zhang2025framepack,cai2025moc,wu2025memorypack}, split the cache into sink, archive, and working banks~\cite{zhang2026memoryforcing}, or retrieve earlier frames by the current prompt~\cite{ji2025memflow}.
In all of them, memory lives inside the generator and is built from its own frames (closed loop).
Our memory lives outside the generator and is written by an agent from the prompts alone.

\paragraph{Agents and memory for multi-shot video.}
Agentic storytelling systems write scripts, draw keyframes, and repair shots with VLM feedback~\cite{wang2026storyengine,huang2026vimax,wu2025movieagent,zheng2024vgot,hu2024storyagent,shi2025animaker,xu2025filmagent,wang2025mavis,xie2024dreamfactory}.
The closest, StoryEngine~\cite{wang2026storyengine}, keeps structured entity states and canonical references.
All of them run offline: they see the full script and can select or regenerate shots.
Concurrent works bring agents closer to streaming.
AgiWorld~\cite{fang2026agiworld} links a planning agent and a video world model through a shared spatial memory.
Wizard~\cite{wu2026wizard} controls a real-time game world model with a VLM agent that turns instructions into programs over scene geometry.
AgiWorld does not report real-time operation, and Wizard's agent edits scene geometry and a symbolic game state, not image memory.
To our knowledge, ours is the first real-time video generator with text and image memory kept by an LLM agent: the video never waits for the agent, and the whole system streams faster than playback.
Other methods keep memory inside the model.
They carry keyframes or long context across shots~\cite{zhang2025storymem,meng2025holocine,guo2025lct,xiao2025captaincinema} or index past frames by pose and time~\cite{xiao2025worldmem,yu2025contextasmemory,li2025vmem}, so their memory comes from generated pixels.
Reference-to-video models~\cite{liu2025phantom,fei2025skyreelsa2,deng2025magref,minimax2026h3} inject a fixed set of reference images.
We instead decide what to store and update it during the session, and our memory images can serve as references.

\paragraph{Asynchronous agents and learned memory.}
Robot systems run a slow planner beside a fast controller~\cite{black2025rtc,shukor2025smolvla,nvidia2025gr00t,shi2025hirobot}, and LLM agents overlap thinking with waiting~\cite{choi2026idlespec,chi2025asyncthink}.
Our slow agent instead must commit versioned memory to a separate real-time generator before a deadline.
LLM agents also learn what to store~\cite{packer2023memgpt,xu2025amem,chhikara2025mem0,zhou2025mem1,yan2025memoryr1,yu2026memagent}, and RL tunes prompts for image and video models~\cite{hao2023promptist,wu2025reprompt,ji2024promptavideo,cheng2025vpo} with learned judges~\cite{liu2025videoalign,xu2024visionreward,he2024videoscore}.
We combine the two.
One LoRA-tuned planner writes the structured memory and the text that the frozen generator reads, and its image reward is calibrated on real videos.

\section{Asynchronous System and Speed}
\label{sec:supp-speed}

\paragraph{Measurement.}
\emph{DiT FPS} is the number of frames divided by denoising time, excluding VAE decoding, as in LongLive-2.0~\cite{chen2026longlive2}.
\emph{End-to-end FPS} divides frames by wall-clock time including decoding, memory prefill, shot-boundary swaps, and waiting. The real-time factor is video seconds over wall-clock seconds.
\emph{Overhead} is the relative increase in wall-clock time over the base generator on the same hardware, prompts, and video length.
Each method is warmed up and loaded once. Events (start and end of each DiT chunk, end of each decode, memory commit, prefill done, boundary swap) are timestamped after \texttt{cuda.synchronize}, and all numbers come from one timing log per run.
DiT throughput varies by $\pm$1--2\% across GPUs, so overheads within $\pm$2\% are treated as noise.
The H100 timing runs on LongLive-2.0 have no entity cards, which the agent writes off the critical path.

\paragraph{Timing runs and quality runs.}
Speed and quality are measured in separate runs. The quality runs generate many stories, seeds, and arms in parallel on H100 and A100 GPUs, so each LongLive-H3 video uses one GPU, and each RAVEN video its official four (\cref{sec:supp-memory}).
The MiniMax-H3 times of \cref{tab:main} come from one session on one H100 node with four GPUs per video (LongLive-H3 with tensor parallelism), on two stories (20 shots) after a one-shot warm-up, with the overhead paired per shot.
GPU count, parallel layout, and faster kernels keep the model, the sampler, and the random draws, and change only BF16 rounding. In blind pairwise judgments, the optimized runtimes do not differ significantly from the release (\cref{tab:supp-speedq}), and neither does the same code on B200 instead of H100.

\begin{table*}[t]
\centering
\caption{Speed configurations behind \cref{tab:speed}. \emph{Time}: seconds per shot (10.54\,s of video on LongLive-2.0, 8\,s on LongLive-H3), including decoding and memory. \emph{RT}: real-time factor. \emph{DiT}: DiT (LongLive-H3: rollout) seconds per shot. \emph{GPUs}: DiT + auxiliary + agent (above eight, on a second node). The LongLive-H3 release rows ran the text encoder on four more GPUs. \emph{Overlapped decoding}: the last chunk of a shot is decoded while the next shot starts.}
\label{tab:supp-speedopt}
\setlength{\tabcolsep}{3pt}
\footnotesize
\resizebox{\linewidth}{!}{%
\begin{tabular}{l ccccl}
\toprule
Configuration & GPUs & Time (s) & RT & DiT (s) & Numerics, lag \\
\midrule
\multicolumn{6}{l}{\textit{LongLive-2.0, H100}} \\
Release inference: BF16, FlashAttention-2, streaming VAE decoder on its own GPU & 1+1+0 & 15.98 & 0.66 & 14.15 & reference \\
FlashAttention-4 + compiled VAE decoder + overlapped decoding & 1+1+0 & 11.63 & 0.91 & 11.47 & kernel order \\
\midrule
\multicolumn{6}{l}{\textit{LongLive-2.0, B200}} \\
Release inference & 1+1+0 & 11.65 & 0.90 & 10.45 & reference \\
cuDNN attention + compiled VAE decoder + overlapped decoding & 1+1+0 & 7.43 & 1.42 & 7.33 & kernel order \\
\rowcolor{oursrow} Full system, live agent, BF16 cuDNN path, prompts at playback pace & 1+2+3 & 7.37 & 1.43 & 7.23 & lag 0 on 40/40 \\
\midrule
\multicolumn{6}{l}{\textit{LongLive-H3, H100}} \\
Release TP inference: 4-GPU tensor parallelism, SDPA, FP32 all-reduce, decoding after the rollout on one GPU & 4+0+3 & 28.80 & 0.28 & 14.25 & reference \\
Fast kernels, 7-GPU tensor parallelism, decoder streamed on an eighth GPU & 7+1+3 & 7.14 & 1.12 & 5.74 & kernel order \\
\midrule
\multicolumn{6}{l}{\textit{LongLive-H3, B200}} \\
Release TP inference & 4+0+3 & 19.94 & 0.40 & 9.77 & reference \\
\rowcolor{oursrow} Fast kernels, decoder streamed on a fifth GPU & 4+1+3 & 5.53 & 1.45 & 4.65 & kernel order \\
\bottomrule
\end{tabular}}
\end{table*}

\paragraph{Deployed agent.}
Qwen3.6-35B-A3B-FP8~\cite{qwen2026qwen36} with multi-token prediction~\cite{deepseek2024v3} on SGLang~\cite{zheng2024sglang} checks images (0.85\,s) and routes scenes (0.06\,s), and FLUX.2-klein-4B~\cite{bfl2026flux2klein} draws (1.2--2\,s), each on one H100.
The RL planner, Qwen3-8B with LoRA served with EAGLE-3~\cite{li2025eagle3} on a third H100, takes 3.5\,s per update in a single live run. Without RL, Qwen3.6-35B also plans.
We chose these models on recorded agent requests. FLUX.2-klein-4B scores below GPT Image but is more than ten times faster, and none of the other fast open image models we compared is better under both of our judges. An API agent (Gemini and GPT Image) needs about 56\,s per update and lags 3.6 shots on LongLive-2.0, so it serves only the offline variants, whose shots take minutes.
An image passes its check if the checker accepts it, or if it scores $\ge$6/10 and shows the required persistent state.
Decoding and prefill use one auxiliary GPU each. Over a 40-shot run, the time per shot stays at 16.0--16.2\,s.

\paragraph{B200.}
On one 8$\times$B200 node (180\,GB each), we keep the story, harness, and GPU layout of \cref{tab:speed} (DiT 1, auxiliary 2, agent 3) and reveal each prompt at playback pace (every 10.54\,s), as a live user would, instead of when the previous shot finishes.
Each cuDNN attention call matches FlashAttention-2 to one or two BF16 rounding steps, but whole videos diverge as with any kernel change, so all quality numbers on LongLive-2.0 and on the chunked-AR MiniMax-H3 generators use the evaluated BF16 path (the bidirectional generators of \cref{tab:supp-bidir-live} run in FP8).
With BF16 weights, cuDNN attention, a compiled VAE decoder, and decoding overlapped across shots, LongLive-2.0 streams at 34.1\,FPS (1.42$\times$) and the full system at 34.3\,FPS with every update in time (\cref{tab:supp-speedopt}). Memory never stalls the video (wait 0.00\,s).
The agent is about 1.6$\times$ faster than on H100 (RL plan 2.3--3\,s instead of 3.5\,s, one image 0.8\,s instead of 1.7\,s). In this live run, version 0 is ready 5.7\,s after a prompt on average (p95 7.8\,s, against 10.54\,s). With a stand-in generator at the same pace, an agent on H100 GPUs has 34 of 39 updates ready in time (\cref{tab:supp-h3lag}). A fourth agent GPU is an option for longer prompts.

\paragraph{Speed optimization.}
\Cref{tab:supp-speedopt} lists the configurations behind \cref{tab:speed}, measured in one session per GPU type on the shots of the timing runs (LongLive-2.0: shots 2--40 of the 40-shot story, LongLive-H3: the 20 full-system shots of the MiniMax-H3 timing runs, after a one-shot warm-up). Its release times are within 1\% of \cref{tab:main}.
\Cref{tab:supp-speedq} checks quality on the same prompts and seeds with a blind pairwise judge (Gemini 3.8 Flash sees two frames of each shot from both versions in random order and prefers one on detail, realism, and overall, with a sign test over shots) and with no-reference metrics. No optimized path differs significantly from the release under the judge. MUSIQ and sharpness of the final LongLive-2.0 path are 3--7\% lower, a shift of the size these metrics also show between paths that the judge cannot tell apart.
\emph{LongLive-2.0:} attention dominates the DiT, so FlashAttention-4 (H100) or cuDNN attention (B200) gives most of the gain. The release decodes the last chunk of each shot after the DiT finishes. Decoding it while the next shot starts removes this tail without changing a bit, and a compiled decoder keeps the decoder GPU ahead of the DiT. One H100 stays below real time even so.
\emph{LongLive-H3:} half of the release time is decoding after the rollout. The fast runtime keeps the model, sampler, and random draws and changes only the kernels (FlashAttention-4 on the KV cache's layout, a fused QKV GEMM, compiled elementwise operations, BF16 all-reduce with sequence parallelism, and no per-step host synchronization). Decoding runs clip by clip as soon as its latents exist, and a fifth GPU for the decoder streams the first frames 1.5\,s after the prompt instead of at the end of the shot.
Memory adds 0.5--0.7\,s per shot, mostly encoding its pictures, which the agent could do when it commits them (\emph{encoded at commit}). Per frame, LongLive-H3 needs 3.9$\times$ the computation of Wan2.2-5B (83 \vs 21\,TFLOP), so it is real time only with more GPUs.
\emph{Agent at H3's pace:} \cref{tab:supp-h3lag} drives the live agent with a stand-in generator at the measured shot times. With the agent on B200 GPUs, every update is ready at the 8\,s playback pace. Without pacing, faster generation leaves less time and one-shot lag appears, so prompts are revealed at playback pace. With the agent on H100 GPUs, version 0 takes 8--9\,s, so most updates arrive one shot late.

\begin{table}[ht]
\centering
\caption{Quality of the speed optimizations: optimized \vs reference on the same prompts and seeds (40 shots on LongLive-2.0, 20 on LongLive-H3). \emph{Judge}: share of shots where the blind pairwise judge prefers each version (optimized / reference), with the sign-test $p$. MUSIQ (no-reference quality) and sharpness (Laplacian): relative difference. \sig: 95\% interval excludes zero.}
\label{tab:supp-speedq}
\setlength{\tabcolsep}{2pt}
\resizebox{\columnwidth}{!}{%
\begin{tabular}{l cccc}
\toprule
Optimized \vs reference & Judge (\%) & $p$ & MUSIQ & Sharpness \\
\midrule
\multicolumn{5}{l}{\textit{LongLive-2.0, B200, \vs BF16 release}} \\
cuDNN + compiled VAE + overlapped decoding & 62.5 / 37.5 & 0.15 & $-$3.3\%\sig & $-$7.2\%\sig \\
\multicolumn{5}{l}{\textit{LongLive-H3, \vs release TP inference}} \\
Fast, 7-GPU TP, full system, H100 / B200 & 45 / 55, 50 / 50 & 0.82, 1.00 & +0.3\%, $-$0.8\% & +3.8\%, $-$4.2\% \\
Fast, 7-GPU TP, no memory, H100 / B200 & 45 / 55, 40 / 60 & 0.82, 0.50 & $-$0.7\%, $-$0.4\% & $-$3.3\%, +0.6\% \\
\bottomrule
\end{tabular}}
\end{table}

\begin{table}[ht]
\centering
\caption{The live agent at the shot times of the optimized generators (stand-in generator, 40 shots). \emph{Budget}: seconds per update. v0: version 0 of the memory ready (mean / p95). \emph{In time}: updates ready before the next shot.}
\label{tab:supp-h3lag}
\setlength{\tabcolsep}{2pt}
\resizebox{\columnwidth}{!}{%
\begin{tabular}{ll cccc}
\toprule
Agent & Generator & Budget (s) & v0 (s) & In time & Lag 0 \\
\midrule
\rowcolor{oursrow} B200 & LongLive-H3, 5 B200s (5.53\,s), paced & 8.00 & 5.8 / 7.8 & 39/39 & 40/40 \\
B200 & LongLive-H3, 5 B200s, not paced & 5.56 & 5.2 / 6.1 & 27/39 & 28/40 \\
B200 & LongLive-H3, 8 B200s, encoded at commit (4.45\,s), not paced & 4.48 & 5.6 / 8.2 & 8/38 & 9/40 \\
B200 & LongLive-2.0, B200 (8.0\,s), paced & 10.54 & 4.7 / 6.6 & 39/39 & 40/40 \\
H100 & LongLive-H3, 8 H100s, encoded at commit (6.82\,s), paced & 8.00 & 8.1 / 9.8 & 12/39 & 13/40 \\
H100 & LongLive-H3, 5 B200s (5.53\,s), paced & 8.00 & 9.0 / 12.8 & 10/39 & 11/40 \\
H100 & LongLive-2.0, B200 (8.0\,s), paced & 10.54 & 8.2 / 12.0 & 34/39 & 35/40 \\
\bottomrule
\end{tabular}}
\end{table}

\paragraph{Bidirectional few-step generators.}
Few-step distillations of MiniMax-H3 that keep its bidirectional attention generate a whole 8\,s shot at once, so a shot is shown about 7\,s after its prompt instead of 1.5\,s for the streamed LongLive-H3. With the same inputs, a blind judge prefers FastH3-Omni over the 50-step model on 71 of 108 shots.
We timed two public ones with reference input, FastH3-Omni with 8 steps~\cite{fastvideo2026fasth3} and Pika-H3-Flex with 6 steps~\cite{pika2026h3flex}, on one 8$\times$B200 node with the requests of the offline H3 variant (2--6 references per shot). Both need FP8 weights and half-size references to stay near 8\,s per shot (BF16 with full-size references: 12.4 and 11.4\,s).
Their sequence parallelism runs on 4, 7, or 8 GPUs (56 attention heads), so the generator takes the whole node, and the agent runs on three GPUs of a second node, which exchanges prompts, pictures, and videos with it over the network (about 0.05\,s per shot).
With our memory, both generators keep pace with prompts revealed every 8\,s (\cref{tab:supp-bidir-live}). With at most three memory pictures, a shot takes 6.6\,s (FastH3-Omni) and 7.5\,s (Pika-H3-Flex), and 38--40 of 40 shots use the newest memory. A shot above 8\,s at the client does not delay the next one, as the generator then waits less for its prompt.

\begin{table}[ht]
\centering
\caption{Live co-runs with bidirectional few-step generators (40-shot story, prompts every 8\,s). The generator takes one 8$\times$B200 node (FP8, half-size pictures) and the agent three B200s of a second node. \emph{Gen.}: generator seconds per shot, mean (max). \emph{Client}: seconds from the prompt to the decoded video, mean (shots within 8\,s). v0, in time, and lag 0 as in \cref{tab:supp-h3lag}.}
\label{tab:supp-bidir-live}
\setlength{\tabcolsep}{2pt}
\resizebox{\columnwidth}{!}{%
\begin{tabular}{ll ccccc}
\toprule
Generator & Injection & Gen. (s) & Client (s) & v0 (s) & In time & Lag 0 \\
\midrule
\rowcolor{oursrow} FastH3-Omni, 8 steps & sink & 6.63 (7.45) & 7.71 (36/40) & 5.2 / 7.0 & 39/39 & 40/40 \\
FastH3-Omni, 8 steps & references & 6.66 (7.84) & 7.70 (37/40) & 5.1 / 8.2 & 37/39 & 38/40 \\
Pika-H3-Flex, 6 steps & sink & 7.54 (7.67) & 7.61 (40/40) & 5.5 / 7.1 & 39/39 & 40/40 \\
Pika-H3-Flex, 6 steps & references & 7.47 (7.62) & 7.55 (40/40) & 5.4 / 6.7 & 39/39 & 40/40 \\
\bottomrule
\end{tabular}}
\end{table}

\begin{figure*}[t]
  \centering
  \includegraphics[width=\linewidth]{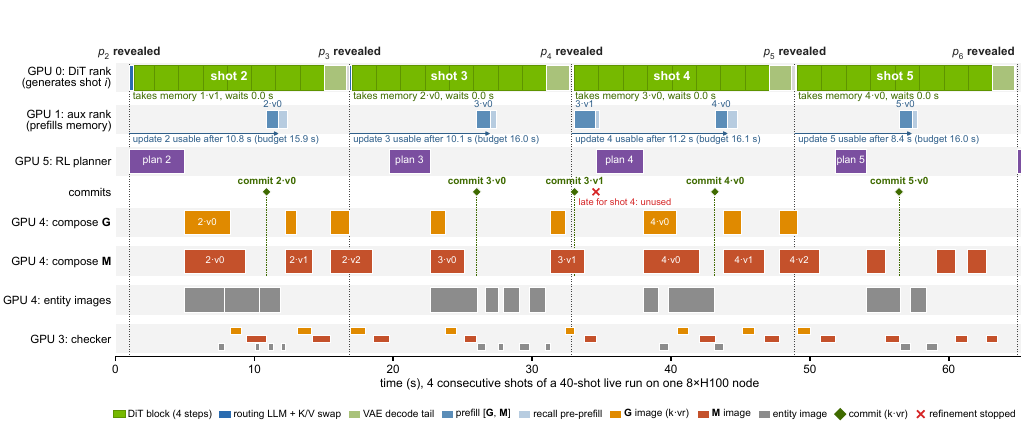}
  \caption{\textbf{A real live run} of the image system on one 8$\times$H100 node (PC-building story, shots 2--5, timestamps from the run's trace). $k\cdot$v$r$: version $r$ of the memory written from prompt $p_k$. A version that arrives after the boundary (red) is not used.}
  \label{fig:supp-gantt}
\end{figure*}

\paragraph{Time budget.}
The agent has one shot to prepare each memory.
When generation is faster than real time, prompts arrive at playback pace, so the budget is the playback time of a shot (10.5\,s for LongLive-2.0 and 8\,s for LongLive-H3 on B200). Otherwise, it is the generation time of a shot (16\,s per 10.5\,s shot for LongLive-2.0 on one H100).
Bounding output tokens and image steps on local open models keeps version 0 within the budget, so agent speed only decides whether refined versions arrive in time.

\paragraph{Refinement.}
Each memory image has up to $R{=}3$ versions, drawn between the reveal of $p_{i-1}$ and that of $p_i$.
Version 0 of $\Mm_i$ is committed as soon as it is drawn, whatever its check says. A later version uses redrawn entity images or appends the checker's complaints to the image prompt, and replaces the committed one only if it passes and scores no lower.
$\Gm$ is replaced only by a passing version, since every later shot reads it.
Each committed version is prefilled on the auxiliary rank at once, and a newer version replaces an older one still waiting in the prefill queue.

\paragraph{A real run.}
\Cref{fig:supp-gantt} shows four shots of a live 40-shot run of the image system (routing + RL planner, as cards add only a string match) on one 8$\times$H100 node, with real timestamps.
Over the run, the DiT rank never waits for memory (at most 0.3\,ms for the lookup), and the routing call and the K/V swap take 0.11\,s per boundary. The RL planner takes 3.5\,s, and version 0 is committed and prefilled 9.8\,s after the prompt on average (at most 13.7\,s), 6.3\,s before the next boundary (at least 2.4\,s).
Later versions replace it only if their check passes. One refined version arrived after the boundary and was simply not used.

\paragraph{Memory versions and lag.}
\Cref{tab:supp-versions-compact} counts, over the 25-story live streams, which version of the newest $\Mm$ each shot receives and how late it is.
Almost every shot gets version 0 (87\% with the 35B planner, 94\% with the RL planner), because refinement is usually cut short by the next prompt. Lag is zero on 99.3\% and 97.3\% of shots, and a refined version reaches 12.8\% (35B) and 5.5\% (RL) of shots, where version 0 failed its check.
Two alternatives cost more on H100 (\cref{sec:speed}): an agent that waits at every boundary for its fully refined memory adds 2.3\% end to end and stalls some shots by up to 2.3\,s, and prefilling on the DiT GPU without an auxiliary rank adds 6.9\%.

\begin{table}[ht]
\centering
\caption{Memory version and lag at each shot boundary in the live streams (25 stories, 16\,s stand-in generator). v$r$: $r$-th refinement round of the newest $\mathbf{M}$ (v0 is committed at once). Lag $k$: memory built from prompts $\le i{-}1{-}k$. RL: three seeds pooled.}
\label{tab:supp-versions-compact}
\setlength{\tabcolsep}{3pt}
\footnotesize
\begin{tabular*}{\columnwidth}{@{\extracolsep{\fill}} ll ccc ccc}
\toprule
& & \multicolumn{3}{c}{$\mathbf{M}$ version (\%)} & \multicolumn{3}{c}{Lag (\%)} \\
\cmidrule(lr){3-5}\cmidrule(lr){6-8}
Planner & Stories & v0 & v1 & v2 & 0 & 1 & $\ge$2 \\
\midrule
35B & All 25 & 87.3 & 12.3 & 0.5 & 99.3 & 0.7 & 0.0 \\
\midrule
\rowcolor{oursrow} RL & All 25 & 94.4 & 5.1 & 0.4 & 97.3 & 2.6 & 0.0 \\
\bottomrule
\end{tabular*}
\end{table}

\section{Memory System}
\label{sec:supp-memory}

\begin{table*}[t]
\centering
\caption{Rows omitted from \cref{tab:main} (same protocol and columns, green: ours). \emph{Image memory}: LLM scene routing and the 35B planner, no cards. \emph{Whole $\Gm$}: the whole $\Gm$ enters the sink instead of its foreground (\cref{tab:supp-fgg}). \emph{Reference}: the same pictures as reference input of H3 Ref2VA, MiniMax-H3's reference-to-video-and-audio model. Gray: entity images as references, seed 0 (\emph{Integration with LongLive-H3}). \emph{Time} on MiniMax-H3: the H100 timing runs of \cref{tab:main}.}
\label{tab:supp-mainrows}
\setlength{\tabcolsep}{4.6pt}
\renewcommand{\arraystretch}{0.92}
\footnotesize
\begin{tabular*}{\linewidth}{@{\extracolsep{\fill}} l ccccc cc l}
\toprule
Method & State & Identity & Same prot. & Location & Revisit & Imaging & Artifacts\down & Time / shot  \\
\midrule
\multicolumn{9}{l}{\textit{LongLive-2.0} \quad (25 stories, 857 shots, 3 seeds)} \\
LongLive-2.0 (built-in memory) & 38.2 & 44.1 & 18.1 & 68.8 & 15.0 & 0.667 & 7.8 & 15.8  \\
\ours \quad Image memory (keyword scene gate) & 45.7 & 60.0 & 21.9 & 78.7 & 17.7 & 0.711 & 6.7 & 15.9 (+0.4\%)  \\
\ours \quad + LLM scene routing & 47.0 & 62.7 & 22.6 & 96.4 & 41.1 & 0.711 & 6.9 & 16.0 (+1.2\%)  \\
\ours \quad + RL planner (image system) & 50.6 & 69.7 & 26.8 & 97.1 & 45.4 & 0.711 & 8.8 & 16.0 (+0.9\%)  \\
\ours \quad + entity cards (full system, \cref{tab:main}) & 64.5 & 89.0 & 46.8 & 98.0 & 55.4 & 0.729 & 6.0 & 16.0 (+0.9\%)  \\
\midrule
\multicolumn{9}{l}{\textit{MiniMax-H3} \quad (25 stories, 857 shots, 2 seeds, time on four H100s per video)} \\
RAVEN (independent shots, \cref{tab:main}) & 36.1 & 61.0 & 24.5 & 84.9 & 26.5 & 0.700 & 0.5 & 44.1  \\
\ours \quad Image memory, whole $\Gm$ & 59.1 & 93.9 & 58.8 & 55.4 & 16.8 & 0.749 & 5.6 & --  \\
\ours \quad Image memory & 59.7 & 94.6 & 59.1 & 88.7 & 41.4 & 0.748 & 1.6 & --  \\
\ours \quad Full system, whole $\Gm$ & 62.5 & 98.1 & 69.4 & 62.6 & 29.5 & 0.753 & 6.2 & 45.3 (+2.6\%)  \\
\arrayrulecolor{black!25}\specialrule{0.3pt}{0.3ex}{0.3ex}\arrayrulecolor{black}
LongLive-H3 (independent shots, \cref{tab:main}) & 35.0 & 58.2 & 20.9 & 85.6 & 24.8 & 0.676 & 0.9 & 28.1  \\
\ours \quad Image memory, whole $\Gm$ & 57.0 & 92.5 & 51.0 & 62.1 & 21.5 & 0.736 & 7.2 & --  \\
\ours \quad Image memory & 58.6 & 92.4 & 51.3 & 84.7 & 36.1 & 0.733 & 4.2 & --  \\
\ours \quad Full system, whole $\Gm$ & 61.2 & 96.5 & 63.0 & 67.4 & 30.6 & 0.740 & 6.1 & 29.1 (+3.9\%)  \\
\ours \quad Image memory (reference), whole $\Gm$ & 61.8 & 94.6 & 47.7 & 65.5 & 50.4 & 0.757 & 12.4 & --  \\
\ours \quad Full system (reference), whole $\Gm$ & 64.5 & 99.5 & 64.4 & 63.8 & 47.6 & 0.761 & 10.3 & --  \\
\gr{\quad Entity images as references} & \gr{62.5} & \gr{95.6} & \gr{44.3} & \gr{73.9} & \gr{50.4} & \gr{0.748} & \gr{15.4} & \gr{--}  \\
\gr{\quad\quad Foreground tokens of the entity images} & \gr{62.1} & \gr{95.9} & \gr{44.1} & \gr{83.5} & \gr{52.7} & \gr{0.746} & \gr{12.4} & \gr{--}  \\
\bottomrule
\end{tabular*}
\end{table*}

\paragraph{Rows omitted from the main table.}
\Cref{tab:supp-mainrows} lists, with all columns of \cref{tab:main}, the rows that the main table omits: the parts of our system added one at a time on LongLive-2.0 (the same runs as \cref{tab:ablation}), image memory without cards and the full system with the whole $\Gm$ on both MiniMax-H3 generators, memory as reference input of H3 Ref2VA, and the agent's entity images as references.
Image memory alone mainly fixes identity on the MiniMax-H3 generators (with foreground $\Gm$, from 61.0\% to 94.6\% on RAVEN and from 58.2\% to 92.4\% on LongLive-H3).
With the whole $\Gm$, both generators copy its plain backdrop at new places and lose location (\emph{Foreground $\Gm$} below).
As reference input of H3 Ref2VA, the same memory follows recalled places better than sinks (revisit 47.6 \vs 30.6 for the full system with the whole $\Gm$) but has more severe artifacts (10.3\% \vs 6.1\%). Entity images as references are discussed in \emph{Integration with LongLive-H3}.

\paragraph{Memory design ablation.}
\Cref{tab:supp-ablation} gives the paired difference of each row of \cref{tab:ablation} to image memory with the 35B planner, and one row that the main table omits.
The \emph{static identity image} is a $\Gm$ of the cast drawn once from the story setup and never updated, without $\Mm$ or scene routing.
\emph{Cards with the 35B planner.} With cards for the entities named in the prompt, the 35B planner reaches 63.9\% state and 87.2\% identity. The full system, which uses the RL planner, is ahead of it by $+0.5$ state, $+1.8$ identity, and $+3.2$ revisit points, none significant, so with cards the RL-trained 8B model can replace the 35B planner.
With the same 35B planner, cards for the named entities add 2.2\sig{} location and 7.2\sig{} revisit points over cards for all injected members, whose long text dilutes place words.

\begin{table}[ht]
\centering
\caption{Paired differences behind \cref{tab:ablation} (LongLive-2.0, 25 stories, 3 seeds): each design minus image memory with the 35B planner, in points ($^\ast$: story-cluster bootstrap 95\% interval excludes zero). The row marked $\circ$ is not in \cref{tab:ablation} (63.9 state, 87.2 identity, 98.2 location, 52.3 revisit).}
\label{tab:supp-ablation}
\setlength{\tabcolsep}{4pt}
\footnotesize
\begin{tabular}{@{}l cc@{}}
\toprule
Design & $\Delta$State & $\Delta$Ident. \\
\midrule
LongLive-2.0, built-in & $-$8.8\sig & $-$18.6\sig \\
\midrule
\multicolumn{3}{@{}l}{\textit{35B planner}} \\
Static identity image only & $-$8.2\sig & $-$12.9\sig \\
$\Gm$ only & $-$6.9\sig & $-$16.6\sig \\
$\Gm$ + newest $\Mm$ & $-$1.4\sig & $-$4.8\sig \\
Keep the last $\Mm$ & $+$0.5 & $+$2.4 \\
Keyword gate & $-$1.4\sig & $-$2.7\sig \\
Cards only (all injected) & $+$12.0\sig & $+$23.8\sig \\
Image mem.\ + all-injected cards & $+$16.4\sig & $+$22.6\sig \\
$\circ$ Image mem.\ + named cards & $+$16.9\sig & $+$24.5\sig \\
\midrule
\multicolumn{3}{@{}l}{\textit{RL planner}} \\
Image memory & $+$3.6\sig & $+$7.0\sig \\
\rowcolor{oursrow} Image mem.\ + named cards (full) & $+$17.4\sig & $+$26.3\sig \\
\bottomrule
\end{tabular}
\end{table}

\paragraph{Scene routing.}
The router is the agent's LLM (Qwen3.6-35B-A3B-FP8, thinking off, temperature 0).
Its input lists the locations of the committed $\Mm$ images, each with the name and description the agent wrote for that scene entity from prompts up to $p_{i-1}$, the location of $\Mm_{i-1}$, and the full prompt $p_i$. The output is constrained to one of the names or NEW.
The system prompt reads: \emph{``For every location that earlier shots were filmed in, the memory keeps a picture. \ldots\ If it takes place in one of the known locations, whether it continues in the previous shot's location or returns to an earlier one, answer with the exact name of that location. If it takes place in a location that is not in the list, answer NEW. Judge by the physical setting the camera shows, not by which characters or objects are mentioned.''}
It was written once and not changed after seeing any decision, and nothing in the router is fitted to \bench.
A recalled $\Mm$ is written right after $\Gm$, attends to $\Gm$, uses its source prompt as text, and the video starts 8 frames later, as for a regular $\Mm$.
On an H100 the call takes 0.06--0.12\,s.
The auxiliary rank keeps $[\Gm, \Mm_{\text{loc}}]$ prefilled for every known location (at most eight), so a recall only swaps blocks. This gives bit-identical video to prefilling at the boundary, which would cost 0.87\,s per recall.
The router keeps $\Mm_{i-1}$ on 69\% of shots, recalls on 17\%, and drops on 14\%.
Against the hidden annotations of the 832 shot boundaries of \bench, it injects an $\Mm$ of the wrong place 19 times (scene gate: 84), an $\Mm$ of the right place 696 times (gate: 568), and recalls the right place on 94\% of the returns whose location is in memory.
The keyword \emph{scene gate} of \cref{tab:ablation} kept $\Mm_{i-1}$ when the opening sentence named its location or shared 20\% of content words with the opening of $p_{i-1}$, and otherwise injected only $\Gm$. Shared words such as ``back'' or ``still'' made it keep the old place on many returns and changes.

\paragraph{H3 prompts.}
MiniMax-H3 generates audio and video from a sectioned prompt.
Before any run, each terse prompt is converted once into this format by Gemini 3.8 Flash from that prompt alone and a few story-level constants (the protagonist's name and voice, the music): one continuous 8\,s shot with an opening phrase (shot size, place, where the protagonist is), the action in timeline order with 0--2 short spoken lines, a soundscape, and the music.
The conversion keeps every fact of the terse prompt and adds no appearance, clothing, colour, layout, or prop detail, so the H3 prompts stay as terse as the originals.
All H3 rows, with and without memory, use the same H3 prompts, and the routing decisions of the H3 rows are made on them.

\paragraph{LongLive-H3.}
We obtain LongLive-H3 from MiniMax-H3 by causal distribution matching distillation (DMD)~\cite{yin2024dmd}, with the chunking and few-step schedule of RAVEN~\cite{lu2026raven}.
The frozen, bidirectional MiniMax-H3 serves as the real-score teacher.
The student is a LoRA of rank 128 on the frozen backbone, and a second rank-128 LoRA serves as the fake-score model, trained by flow matching on the student's own rollouts. The student is trained with the normalized DMD surrogate, and fake-score and student updates alternate 5:1 after 50 fake-score-only warm-up updates.
Training uses prompts only, without real video or audio: 1{,}024 original single-shot prompts from SMD-H3-T2VA-100K and 256 multi-shot prompts that we rebuilt from SMD.
Training clips have 124 frames at 1344$\times$768 and 24\,fps with jointly generated 32\,kHz stereo audio, \ie 37 video latents and 207 audio latents per channel. Each clip is split into eight chunks, seven of 5 video latents (about 0.7\,s) and a 2-latent tail, with the audio split at the same boundaries.
Attention is block-causal: each chunk attends to the text, the first chunk as a sink, the two previous chunks, and itself, and earlier chunks are fed as clean context at clock 0.999.
We train on 32 GB300 GPUs with a global batch of 64 (2 per GPU), learning rates of $2\times10^{-6}$ for the student and $10^{-5}$ for the fake-score model, and EMA decay 0.99. The released checkpoint is the EMA after 200 student updates (1{,}000 fake-score updates).
At inference, LongLive-H3 takes 3 steps per chunk and generates 8\,s shots (192 frames), longer than its training clips.
With a KV cache that reproduces its chunked sampler, a shot takes 50.2\,s on one H100 (116\,s on A100), and all quality results use this one-GPU path. \Cref{tab:main} times it on four H100s with tensor parallelism (\cref{sec:supp-speed}).
RAVEN generates 1376$\times$768 as released, with 4 steps per chunk.

\paragraph{K/V carried across shots.}
Without memory, every MiniMax-H3 shot is a separate rollout from its prompt. The second baseline carries K/V across shots as the built-in memory of LongLive-2.0 does, with the mechanism of our sink: the first latent frames of the story's first shot (a global sink) and the last latent frames of the previous shot are written as one clean chunk after the new shot's text, so their K/V are recomputed under the new prompt (LongLive's K/V re-cache~\cite{yang2025longlive}). RAVEN carries four global and six recent frames (1\,s, its K/V window). LongLive-H3, which holds the whole model on one 80\,GB GPU in the quality runs, carries one global and two recent frames, the size of our memory chunk. The carried frames end eight video frames (1/3\,s) before the new shot, as in LongLive-2.0's multi-shot format, the shot starts at 2\,s, as with our memory chunk, and prompts, seeds, and everything else equal the rows without memory.
Carried K/V raises identity over no memory (RAVEN 61.0\% $\to$ 91.2\%, LongLive-H3 58.2\% $\to$ 76.0\%, both \sig), though less than our full system (96.8--98.0\%), but shots continue the previous one: location on scene changes falls to 20--26\% (85--86\% without memory), and revisits do not improve.
Severe artifacts also accumulate along the story (RAVEN 1.6\% of shots 1--10 and 14.9\% after shot 20, LongLive-H3 3.6\% and 17.5\%), mostly colour distortion or corrupted frames, as each shot continues frames that the generator produced under another prompt. With our memory, which comes from the agent rather than from the video, they stay flat along the story (3--6\%).

\paragraph{Integration with LongLive-H3.}
All arms use the same H3 prompts, the same two video seeds, and the same memory stream of the 35B planner, routed by the same LLM router on the H3 prompts (the right $\Mm$ on 642 of 832 boundaries, against 696 on the terse prompts). Since H3 places text at the start of the sequence, memory K/V depend on the prompt and are computed after it arrives (\cref{sec:speed}). We compare three ways to pass memory.
\emph{Sink:} on keep, $[\Gm, \Mm_{i-2}, \Mm_{i-1}]$, on recall, $[\Gm, \Mm_{\text{loc}}]$, and on drop, $[\Gm]$, as one clean chunk after the text at 0, 0.25, and 0.5\,s, with the shot starting at 2\,s.
\emph{Reference:} the H3 Ref2VA transformer with LongLive-H3's weight update receives the same routed pictures as references, and a fixed template wraps the shot prompt in H3's reference-prompt sections, with no extra LLM call.
\emph{Entity references:} the same transformer receives up to six committed entity images (one of them the newest plate of the routed place) and the shot prompt rewritten by Qwen3.6-35B-A3B into H3's full-reference format, without $\Gm$ or $\Mm$. The rewrite takes 6.7\,s on one H100 after the prompt, which C3 rules out.
Against our full system (seed 0), entity references reach the same state and identity and a higher revisit ($+8.2$\sig), but lower same protagonist ($-20.3$\sig), location ($-13.8$\sig), and framing ($-21.8$\sig) and 11.2\sig{} points more severe artifacts, as shots often open on one of the pictures and then jump into the scene. Passing only the foreground tokens of each entity view brings location back to 83.5\% but leaves severe artifacts at 12.4\%. Sink memory instead composes the agent's choice into two prefilled images, whose content the RL planner learns.

\paragraph{Runs on MiniMax-H3.}
RAVEN uses its official configuration and weights (the MiniMax-H3 base model with a 4-step causal streaming LoRA and a window of 2 sink and 2 recent chunks) on four GPUs per video, with the same H3 prompts, memory stream, routing decisions, and sink layout as LongLive-H3, so the two sinks differ only in the backbone.
The full-system rows (sinks in \cref{tab:main,tab:supp-mainrows}, references in \cref{tab:supp-mainrows}) use the memory stream of the RL planner (training seed 0, the stream of the full system on LongLive-2.0) with the same LLM scene router run on the H3 prompts, and append the same entity cards as on LongLive-2.0 at the end of the description section of each H3 prompt, before the sound sections. The image-memory rows use the 35B planner's stream with the same router and no cards. MiniMax-H3 encodes the whole prompt (no 512-token window).
All rows use video seeds 0 and 1000 and are compared with the same seeds of the same backbone without memory, with both arms of a story on one GPU type, so every difference is paired by story and seed.

\paragraph{Bidirectional generators.}
Sinks need no causal attention. MiniMax-H3 packs the text, the condition images, and the generated video into one sequence on a shared time axis (40 positions per second). Condition rows stay clean and only the video is denoised. Ref2VA places each reference one position after the previous one and starts the video right after the last. For a sink, we move the memory pictures to 0, 0.25, and 0.5\,s and the video to 2\,s, as on LongLive-H3, and replace the per-picture prompt lines by one sentence saying that the pictures are frames of earlier shots and that the shot starts after a cut. Rows and computation equal those of Ref2VA references, but the pictures are attended to at every denoising step instead of being prefilled once, and memory can change only between shots.
With the per-picture lines of the LongLive-H3 prompts, Pika-H3-Flex often opens on the memory pictures, and the judge prefers the single sentence on 45 of 54 shots.

\paragraph{Foreground $\Gm$.}
$\Gm$ shows its members on a plain light-grey backdrop, which covers 70\% of its tokens. MiniMax-H3 generators attend to every sink token in every chunk and read this backdrop as a place: with the whole $\Gm$, the judge describes a studio or plain backdrop on 36--52\% of the new-place shots, where routing injects $\Gm$ alone, and on 10--15\% of the shots with $[\Gm, \Mm]$ (seed 0).
We therefore keep only the foreground of $\Gm$. When $\Gm$ is committed, the agent masks it with the union of a salient-object matte (BiRefNet~\cite{zheng2024birefnet}, threshold 0.5) and a colour key for smooth, unsaturated, bright regions connected to the image border. The mask is fitted to the video like $\Gm$ and pooled over the $32{\times}32$-pixel cell of each token, and a token is kept, at its original $(t,h,w)$ position, if at least 10\% of its cell is foreground (30\% of the tokens on average). Backdrop tokens are never written, and the caption of $\Gm$ drops its sentence on the backdrop. The mask takes under 0.5\,s per $\Gm$ before the prompt, and nothing else changes.
With only this change (\cref{tab:supp-fgg}), foreground $\Gm$ gains 20.1\sig{} (LongLive-H3) and 26.4\sig{} (RAVEN) location points over the whole $\Gm$, back to about the level without memory, and 12.2\sig{} and 17.9\sig{} revisit points, while state and identity change by less than 1 point. Studio mentions on new-place shots fall to 6--8\%. Both parts are needed: on the new-place shots of LongLive-H3 (seed 0), dropping only the backdrop sentence gains 1.9 location points (not significant), only the backdrop tokens 5.7\sig, and both 10.7\sig. Image memory without cards gains in the same way (\cref{tab:supp-mainrows}).
LongLive-2.0 does not copy the backdrop. It pins the first chunk of each new shot, generated after ``The scene transitions.'', as the shot sink (\cref{sec:memory}), so later chunks follow the new shot's opening and its prompt: with the whole $\Gm$, all 111 new-place shots that receive $\Gm$ alone reach their planned place (seed 0). Foreground $\Gm$ there changes location, state, identity, revisit, and framing by less than 1 point and lowers same protagonist by 4.6\sig{} points (\cref{tab:supp-fgg}), so LongLive-2.0 keeps the whole $\Gm$.

\begin{table}[ht]
\centering
\caption{Foreground \vs whole $\Gm$ (full system with sinks, paired by story). MiniMax-H3: video seeds 0 and 1000 (50 story runs, the full-system rows of \cref{tab:main}). LongLive-2.0: seed 0 (25 stories), whose rows in \cref{tab:main} keep the whole $\Gm$. $\Delta$: foreground $-$ whole ($^\ast$: story-cluster bootstrap 95\% interval excludes zero).}
\label{tab:supp-fgg}
\setlength{\tabcolsep}{2.2pt}
\resizebox{\columnwidth}{!}{%
\begin{tabular}{@{}l ccc ccccc@{}}
\toprule
& \multicolumn{3}{c}{Location} & \multicolumn{5}{c}{$\Delta$} \\
\cmidrule(lr){2-4}\cmidrule(l){5-9}
Generator & Whole & Fg. & $\Delta$ & State & Ident. & Same & Rev. & Artif. \\
\midrule
LongLive-H3 & 67.4 & 87.5 & $+$20.1\sig & $-$0.0 & $+$0.4 & $-$0.3 & $+$12.2\sig & $-$1.1 \\
RAVEN & 62.6 & 89.1 & $+$26.4\sig & $+$0.8 & $+$0.1 & $+$0.7 & $+$17.9\sig & $-$3.2\sig \\
LongLive-2.0 & 96.9 & 97.3 & $+$0.4 & $+$0.2 & $-$0.2 & $-$4.6\sig & $-$0.9 & $+$2.5 \\
\bottomrule
\end{tabular}}
\end{table}

\paragraph{Entity cards.}
Cards for all injected $\Gm$/$\Mm$ members raise state and identity strongly but lower location and framing, as the long cards dilute place and camera words (\cref{tab:ablation}, framing in \cref{sec:supp-limits}). Cards only for the entities whose \emph{name appears in the revealed prompt} (string matching against the committed entity bank, 2.4 cards per shot instead of 3.9, no LLM call after the prompt, so C3 still holds) keep almost all of the gain with location and framing unchanged. Over the three seeds of \cref{tab:main}, they raise state by 13.9\sig{} and identity by 19.3\sig{} points over the image system.
A card reads, \eg, ``Remembered from earlier shots: Sam: Sam, a tall broad man in his mid-30s with sandy blonde hair. Current state: clothing: maroon t-shirt and grey sweatpants.'' Cards follow the prompt, so the 512-token window of the text encoder, which 0.2\% of prompts exceed, never cuts the user's words. Matching takes under 1\,ms, and text encoding about 0.1\,s per shot with or without cards. Like the images, a card reflects memory up to shot $i{-}1$.
\emph{Naive text history.} To test whether cards help only because the prompt carries more text, a baseline without the agent appends, for each entity named in the prompt, the raw sentences of earlier prompts that mention it (its first mention and the two latest, about 155 words per shot). On LongLive-2.0 alone, this raises identity from 46.5\% to 80.2\% but costs 11.3\sig{} framing points, as old sentences bring back old camera instructions (\cref{tab:supp-hist}). With the same image memory, the agent's cards beat the history by 5.0\sig{} state, 8.8\sig{} framing, and 4.3\sig{} revisit points: they carry the current state that the agent tracked, in a few words, instead of every earlier mention.

\begin{table}[ht]
\centering
\caption{Naive text history \vs entity cards (seed 0, 25 stories, LongLive-2.0). History: raw earlier prompt sentences of each named entity, no agent.}
\label{tab:supp-hist}
\setlength{\tabcolsep}{2.0pt}
\resizebox{\columnwidth}{!}{%
\begin{tabular}{l ccccccc}
\toprule
Design & State & Ident. & Same & Loc. & Fram. & Rev. & Imaging \\
\midrule
LongLive-2.0 (built-in memory) & 38.6 & 46.5 & 19.1 & 62.5 & 57.1 & 13.2 & 0.664 \\
\quad + text history & 51.2 & 80.2 & 41.4 & 69.3 & 45.7 & 28.0 & 0.669 \\
Image system (routing + RL) & 50.2 & 64.3 & 25.6 & 96.9 & \best{66.2} & 44.3 & 0.710 \\
\quad + text history & 58.8 & 86.4 & 43.1 & 93.9 & 57.4 & 51.2 & \best{0.730} \\
\rowcolor{oursrow} \quad + entity cards (full system) & \best{63.8} & \best{89.4} & \best{47.1} & \best{97.3} & \best{66.2} & \best{55.5} & 0.729 \\
\bottomrule
\end{tabular}}
\end{table}

\paragraph{State over 14 minutes.}
On the nine stories extended to 80 shots (two seeds, \cref{fig:analysis}), image memory keeps state ahead of LongLive-2.0 through the whole story: the image system leads by $+9.8$\sig{} and $+11.3$\sig{} state points in the first and last seven minutes, and the full system by $+9.3$\sig{} and $+6.8$.
The full system also leads by 48.6\sig{} and 61.7\sig{} identity points and by 0.067\sig{} and 0.107\sig{} in imaging quality.
Entity cards raise identity over the image system by 18.2\sig{} points (89.3\% \vs 70.9\% in the first seven minutes, 87.8\% \vs 69.7\% in the last seven) and imaging quality by 0.012\sig{} at the same state.
Prompts with cards are as long as on \bench (295 tokens on average, 0.3\% over 512), so the text window does not explain this.

\paragraph{Entity layer.}
The entity layer follows from the online setting.
\emph{Views:} each character and object is drawn in two views on a plain light-grey backdrop, a front view (full body for characters, the whole object for objects) and a three-quarter view. Each scene is drawn once as an empty plate, a wide view of the place without people. The three-quarter view is drawn with the front view as a reference, so both show the same entity. $\Gm$ and $\Mm$ are composed from the front views and the plates only, so on LongLive-2.0 the three-quarter views never reach the generator. They are used only when entity images themselves are passed as references (\emph{Integration with LongLive-H3}), where all views of a selected entity are listed together.
\emph{Persistent versus transient (C1):} entity images show only attributes that persist across shots, such as identity, clothing, dirt, damage, or assembly progress. Pose, action, and held objects stay neutral because the next prompt decides them.
\emph{Memory as current belief (C1, C4):} each state records its source shot and whether it was stated or inferred, and a contradicting later prompt revokes it and triggers a redraw.
\emph{Refinement:} a new entity is drawn from text and a changed one as an edit of its latest image, retried from its first image if the edit fails. Each entity image is checked and repaired at most once, and if it still fails, the entity keeps its previous image. Memory images are refined as in \emph{Refinement} (\cref{sec:supp-speed}).
These choices decide what memory means when the next prompt is unknown, and they matter whenever an image would otherwise show a pose or a revoked state.

\section{Memory Planner}
\label{sec:supp-rl}

\paragraph{Training details.}
The policy is Qwen3-8B with LoRA ($r{=}32$, $\alpha{=}64$, all linear layers) and the deployed system prompt and JSON schema.
Training uses 56 stories plus 8 development stories written by Qwen3.6 (24--32 shots each) whose per-entity persistent states are accumulated by code.
RFT uses 2,865 examples (best of 4 by reward from 3,138 teacher states).
GRPO treats each shot as one decision and runs 80 steps with 8 states $\times$ 8 samples per step, the token-level loss of DAPO~\cite{yu2025dapo}, no KL term, zero-variance groups dropped, temperature 0.9, learning rate $2{\times}10^{-5}$, gradient clipping 0.5, group standard-deviation floor 0.1, and one-step off-policy pipelining. Each seed trains on one H100 and shares rollout, judge, and image services on two 8$\times$H100 nodes (about 2--2.5 hours for three seeds in parallel).
The training judge is Qwen3.6-35B-A3B and the test judge Gemini 3.8 Flash with the same prompts. They agree on 79--83\% of items (per-decision correlation 0.66--0.77).

\paragraph{Reward details.}
In \cref{eq:rimg}, the text judge scores each persistent state that the next six shots need $+1$ if memory holds it correctly, $0$ if it is missing, and $-1.5$ if it is wrong.
The image term renders $\Gm$ and $\Mm$ with a fixed seed. Each entity of the next shot that the VLM finds recognizable and correctly drawn earns 0.51 if it is in $\Gm$ and 0.49 if it is in $\Mm$ (down to half if drawn in a wrong state), and each member of $\Mm$ absent from the next shot costs 0.11.
These weights come from a regression of 3,867 judged entity attributes from real videos on memory composition, with story and planner fixed effects: an on-screen entity gains $+9.0$ points of state accuracy (95\% CI $[4.7, 13.8]$) in $\Gm$ and $+8.7$ $[4.6, 12.4]$ in the latest $\Mm$, while each absent member of $\Mm$ costs $-1.9$ $[-3.7, 0.0]$. Normalizing these effects gives 0.51, 0.49, and 0.11.

\paragraph{All planner columns.}
\Cref{tab:supp-rl} adds the columns that \cref{tab:rl} omits: contradictions in memory and same protagonist in the video.

\begin{table}[ht]
\centering
\caption{Columns that \cref{tab:rl} omits. \emph{Contra.}: share of needed facts that memory contradicts (\%). \emph{Same}: same protagonist in the video, difference to the deployed 35B planner on the same stories and seeds. RL rows average 3 training seeds (text reward only: 2), as in \cref{tab:rl}.}
\label{tab:supp-rl}
\setlength{\tabcolsep}{4pt}
\footnotesize
\begin{tabular}{l cc}
\toprule
Planner & Contra.\down & Same \\
\midrule
Qwen3-8B, frozen & 15.1 & $-$0.9 \\
\quad + SFT (RFT warm start) & 11.9 & $-$0.8 \\
Qwen3.6-35B (deployed) & 9.7 & -- \\
RL, text reward only & 12.2 & $-$6.6\sig \\
RL, calibrated, not rendered & 11.6 & $+$2.2 \\
\rowcolor{oursrow} RL, video-calibrated & 10.4 & $+$4.2\sig \\
\bottomrule
\end{tabular}
\end{table}

\paragraph{Calibration under the deployed $\Gm$.}
The image weights were estimated on development videos whose $\Gm$ also shows the current scene instead of a plain backdrop, and the policy was trained with them.
As a check that is not used for training, we fit the same model on the benchmark videos of the RL streams (16 LongStory-Hard stories, three video seeds), once on the videos made with the earlier $\Gm$ and once on those made with the deployed $\Gm$.
An entity in the deployed $\Gm$ gains as much fact accuracy as in the earlier one ($+7.9$ $[4.1, 11.6]$ \vs $+8.2$ $[3.4, 12.6]$ points), so the calibration still holds where the planner is deployed.

\begin{figure}[t]
\centering
\includegraphics[width=\linewidth]{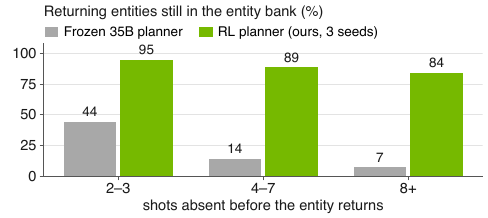}
\caption{\textbf{Returning entities in the entity bank}: share of entities returning after 2--3, 4--7, or 8+ absent shots that the planner has not retired when the return prompt arrives (16 LongStory-Hard stories). The three RL seeds differ by at most 3 points.}
\label{fig:supp-retention}
\end{figure}

\paragraph{Revisits.}
We count the returns of entities after 2--3 / 4--7 / 8+ absent shots in the live streams of the 16 LongStory-Hard stories, whose hidden annotations list the entities of every shot (533 returns).
The share already drawn in the injected $\Gm$ or $\Mm$ of the return shot is 49/13/5\% for the 35B planner and 58/21/11\% for the RL planner.
The share still in the entity bank when the return prompt arrives, \ie, not retired by the planner, is 44/14/7\% for the 35B planner and 95/89/84\% for the RL planner (\cref{fig:supp-retention}).
The RL planner keeps returning entities in the bank and draws them once they return (in the $\Gm$ or $\Mm$ of the next shot for 87/71/67\% of returns, \vs 82/61/54\% for the 35B planner).

\paragraph{Memory before and after RL.}
Over the 605 decision points on LongStory-Hard where both planners have zero lag, the RL planner's $\Gm$ and $\Mm$ hold 69\% of the next shot's returning entities, against 60\% for the 35B planner, and for entities that only the RL planner keeps, the video's state accuracy on their facts is 38.9\% \vs 25.2\%.

\paragraph{Action space.}
The RL planner reads the entity bank as text (each entity's states, the last shot that showed it, and its location).
Its action has typed state fields (entity descriptions, facts with their persistent and source flags, revocations, scene, and the $\Gm$/$\Mm$ members) and free text (image prompts, $\Gm$/$\Mm$ layout prompts, and sink captions), which is 51\% of its output tokens.

\section{Benchmark and Metrics}
\label{sec:supp-bench}

\paragraph{Terse prompts.}
Terse prompts are rewritten from full prompts by Gemini 3.8 Flash. The first shot is unchanged, and new places keep their description, while continuing or revisited places become one phrase (``still in'', ``back in''). The protagonist and earlier entities are named without appearance or state, changes that happen in the shot remain as actions, and camera motion is kept.
Automatic checks reject rewrites that repeat attribute phrases of a character card or whose place part exceeds 15 words for continued places (at most two retries).

\paragraph{LongStory-Hard.}
Sixteen stories with about 40 shots and 13--14 entities each (12--14 recurring, 2--4 per shot), with 7--36 returns after 5+ absent shots per story.
They are written by Gemini 3.8 Flash, a different model from the one that wrote the training stories, on topics disjoint from training, development, and hand-written stories.
Hidden states are accumulated by code from initial attributes and per-shot persistent changes.
The manifest is frozen with a hash before any method is run.

\paragraph{Judges.}
The shot judge scores location, framing, content, identity, persistent states, and artifacts from four frames (4, 35, 65, 97\% of the shot), and the pairwise judge scores same place and same protagonist (1--5, camera changes not penalized). The pairwise metrics are described below.
Re-judging 211 shots agrees on 95.7\% (location), 94.8\% (framing), and 92.3\% (state items).
\paragraph{Second judge.}
Gemini 3.8 Flash also wrote the Hard stories and terse prompts, so we re-judged seed 0 of the 14 real-time runs (all Wan2.1 models, every LongLive-2.0 row including the full system, and five MiniMax-H3 runs, about 12{,}000 shots) with GPT-5.5 (low reasoning effort), with byte-identical judge prompts, frames, and scoring code (\cref{tab:supp-judge2}). These runs use a $\Gm$ that also shows the current scene instead of the plain-backdrop $\Gm$ of \cref{sec:memory}, so their values differ slightly from \cref{tab:main}, but the comparison concerns the agreement of the two judges on the same shots.
Per item, they agree with Cohen's $\kappa$ 0.69--0.80 on state, identity, location, and framing, and with Spearman $\rho$ 0.66--0.73 on the pairwise ratings. Over the 14 runs, their columns correlate at $r = 0.98$--$0.99$.
Of the 94 comparisons that were significant at seed 0 under Gemini, 87 keep their sign and significance under GPT-5.5 and none flips sign, so the gains of the full system and the entity cards do not depend on the judge. Severe artifacts are the exception ($\kappa$ 0.28), so we draw no conclusion from small artifact differences.

\begin{table}[ht]
\centering
\caption{Second judge: seed 0, all 25 stories, Gemini 3.8 Flash $\rightarrow$ GPT-5.5 on the same shots. SkyReels-V2 (not real time) is shown for reference, and the five MiniMax-H3 runs enter only the statistics in the text. --: not available.}
\label{tab:supp-judge2}
\setlength{\tabcolsep}{1.6pt}
\resizebox{\columnwidth}{!}{%
\begin{tabular}{l cccccc}
\toprule
Method & State & Identity & Same prot. & Location & Framing & Revisit \\
\midrule
CausVid & 23.2\,$\rightarrow$\,26.0 & 41.8\,$\rightarrow$\,32.4 & 18.5\,$\rightarrow$\,24.2 & 35.2\,$\rightarrow$\,38.8 & 22.6\,$\rightarrow$\,24.3 & 9.3\,$\rightarrow$\,13.0 \\
SkyReels-V2 & 30.8\,$\rightarrow$\,31.2 & 49.2\,$\rightarrow$\,38.6 & 18.9\,$\rightarrow$\,24.4 & 46.7\,$\rightarrow$\,47.1 & 31.5\,$\rightarrow$\,29.0 & 10.5\,$\rightarrow$\,12.2 \\
Self-Forcing & 35.3\,$\rightarrow$\,30.4 & 54.2\,$\rightarrow$\,37.9 & 11.9\,$\rightarrow$\,23.3 & 9.2\,$\rightarrow$\,18.9 & 9.8\,$\rightarrow$\,12.3 & 4.6\,$\rightarrow$\,9.8 \\
Rolling Forcing & 30.4\,$\rightarrow$\,30.8 & 58.1\,$\rightarrow$\,50.8 & 24.8\,$\rightarrow$\,33.6 & 33.0\,$\rightarrow$\,31.9 & 28.1\,$\rightarrow$\,30.0 & 16.4\,$\rightarrow$\,18.0 \\
Causal-Forcing & 28.4\,$\rightarrow$\,29.0 & 60.4\,$\rightarrow$\,49.3 & -- & 58.6\,$\rightarrow$\,66.4 & 25.3\,$\rightarrow$\,20.1 & 11.6\,$\rightarrow$\,19.7 \\
LongLive-2.0 & 37.8\,$\rightarrow$\,37.1 & 46.0\,$\rightarrow$\,35.6 & 18.2\,$\rightarrow$\,24.5 & 64.4\,$\rightarrow$\,58.2 & 57.6\,$\rightarrow$\,55.0 & 12.5\,$\rightarrow$\,14.6 \\
+ our memory (gate) & 46.5\,$\rightarrow$\,45.6 & 65.1\,$\rightarrow$\,53.8 & 24.6\,$\rightarrow$\,32.7 & 71.3\,$\rightarrow$\,68.3 & 67.3\,$\rightarrow$\,63.0 & 17.0\,$\rightarrow$\,16.4 \\
+ LLM scene routing & 47.0\,$\rightarrow$\,44.6 & 65.9\,$\rightarrow$\,56.5 & 24.2\,$\rightarrow$\,32.8 & 94.6\,$\rightarrow$\,93.4 & 68.6\,$\rightarrow$\,63.4 & 43.6\,$\rightarrow$\,51.6 \\
+ RL planner (image system) & 50.9\,$\rightarrow$\,49.5 & 71.1\,$\rightarrow$\,64.5 & 28.8\,$\rightarrow$\,39.3 & 92.3\,$\rightarrow$\,92.0 & 66.3\,$\rightarrow$\,61.7 & 44.3\,$\rightarrow$\,52.9 \\
\rowcolor{oursrow} + entity cards (full system) & 64.1\,$\rightarrow$\,61.7 & 89.2\,$\rightarrow$\,88.6 & 49.4\,$\rightarrow$\,-- & 92.3\,$\rightarrow$\,90.7 & 65.2\,$\rightarrow$\,66.1 & 55.0\,$\rightarrow$\,67.6 \\
\bottomrule
\end{tabular}}
\end{table}

\paragraph{Benchmark statistics.}
\Cref{tab:supp-benchstats} summarizes \bench.
Each shot's hidden annotation lists its place, planned camera and content, and the expected persistent states that the judge checks. The Hard stories also list the entities present, their first appearances, and their changes.

\begin{table}[ht]
\centering
\caption{\textbf{\bench statistics.} Places: scene changes between consecutive shots (new place, continued place, revisit of an earlier place). Returns: an entity seen again after 5+ absent shots. Prompt words: full \vs terse prompts.}
\label{tab:supp-benchstats}
\setlength{\tabcolsep}{3pt}
\resizebox{\columnwidth}{!}{%
\begin{tabular}{l ccc}
\toprule
& Hand-written & LongStory-Hard & All \\
\midrule
Stories & 9 & 16 & 25 \\
Shots (per story) & 211 (14--40) & 646 (38--42) & 857 \\
Places per story & 7.6 & 4.9 & 5.8 \\
New / continued / revisited & 59 / 131 / 12 & 62 / 440 / 128 & 121 / 571 / 140 \\
Entities per story & -- & 13.8 & -- \\
Entities per shot & -- & 2.8 & -- \\
Returns after 5+ absent shots & -- & 313 & -- \\
State keys per story & 3.2 & 30.4 & 20.6 \\
Judged state items per shot & 3.0 & 8.3 & 7.0 \\
Prompt words (full / terse) & 239 / 156 & 122 / 58 & 164 / 93 \\
\bottomrule
\end{tabular}}
\end{table}

\paragraph{Judge prompts.}
The shot judge receives the system prompt below, then the place list (the setting sentence of the first prompt of each annotated place), the protagonist card, the planned camera and content, the expected states, and the four frames. The pair judge receives its system prompt, the place name, the protagonist card, and two frames of each shot. Both run at temperature 0 with JSON output.
\par\nolinenumbers
\par\noindent\begin{minipage}{\linewidth}
\begin{lstlisting}[title={Shot judge (system prompt)}]
You evaluate one shot of an AI-generated long video against a hidden plan. You see FRAMES of the shot in time order (first frame, about 1/3, about 2/3, last frame). Judge only what is visible. Return JSON:
{"opening_location": str, "location": str, "shot_size": "extreme close-up|close-up|medium|wide",
 "framing_match": bool, "content": 0|1|2, "protagonist_visible": bool, "protagonist_match": bool|null,
 "facts": [{"key": str, "verdict": "confirmed|contradicted|not_visible"}], "artifacts": bool, "notes": str}
- opening_location: where the FIRST frame takes place; location: where most of the shot takes place. Each is exactly one key from LOCATIONS, or "other".
- framing_match: true if the camera distance/angle and the main subject agree with PLANNED CAMERA and PLANNED CONTENT (e.g. a planned close-up of hands on a part must not be a medium shot of the person standing).
- content: 2 if the main subject and action of PLANNED CONTENT are clearly shown, 1 if only partly (right subject or place but the action is missing or different), 0 if not shown.
- protagonist_match: if the PROTAGONIST is visible, whether it matches the identity description (face, species, body, age; ignore clothing and accessories, which may change); null if not visible.
- facts: one entry per EXPECTED STATE key: "confirmed" if clearly visible and correct, "contradicted" if visible and different, "not_visible" otherwise.
- artifacts: true only for severe failures: a ghosted overlay or blend of two different images, melting or broken anatomy, frozen or corrupted frames.
\end{lstlisting}
\end{minipage}

\begin{lstlisting}[title={Pair judge (system prompt)}]
You compare two shots, A and B, of an AI-generated story video. The story says both take place in the same location. You see two frames of shot A, then two frames of shot B. Judge only what is visible; a different camera angle, distance or time of day is expected and must not lower the score. Return JSON:
{"place": 1-5, "person": 1-5 or null, "shared": str, "differs": str}
- place: how consistent the place is. 5 = unmistakably the same place: specific furniture or landmarks, layout, materials and colours match; 4 = the same place with minor inconsistencies; 3 = same kind of place with a few specific matching details but clear differences; 2 = same kind of place, no specific details match; 1 = a different kind of place. If a shot is a close-up, compare whatever background, surfaces and props are visible.
- person: if the PROTAGONIST is visible in both shots, how consistent their identity is (face, hair, build, species and markings; clothing, dirt and accessories may legitimately change). 5 = clearly the same individual, 3 = similar but uncertain, 1 = clearly a different individual; null if not visible in both.
- shared / differs: the main matching and differing details (short).
\end{lstlisting}
\linenumbers

\paragraph{Cross-shot metrics.}
\emph{Same protagonist} fixes the shots that count from the plan: the evaluation judge reads only the hidden plan of each shot (planned camera, content, and entities present), never a video, and decides whether it shows the protagonist recognisably (face, or head and body for a non-human protagonist, but not only hands, from behind, or tiny). This holds for 349 of 857 shots (41\%), with a prompt written once.
Each of these shots scores (rating$-$1)/4 against the method's first shot of the protagonist (four frames of each shot, compared on face, hair, build, age, species, and markings, while clothing, dirt, and pose may change) and 0 if the shot judge does not see the protagonist. The reference shot itself is excluded.
Because the shots are fixed by the plan, a video that loses the protagonist or breaks down cannot score by showing the protagonist less often, and because pairs at any place and distance count, carrying the previous shot's pixels forward does not help either.
\emph{Revisit} rates whether a returning shot shows the same place as the last visit and counts a return only if both shots are judged to be at their planned location, as the geometric place consistency of StoryEngine~\cite{wang2026storyengine} counts only shots in the right scene, so a model that stays in one place cannot score by looking the same.

\paragraph{Face identity (MIR).}
MIR follows StoryEngine exactly (its App.~D.8). It takes four evenly spaced frames per shot, including the first and last, and keeps InsightFace buffalo\_l detections at 640$\times$640 with score $\ge$0.55. In shot order, each embedding joins the cluster with the most similar mean if the cosine is $\ge$0.45 and otherwise starts a new one, and single-detection clusters are dropped. MIR is the minimum over clusters of the mean cosine between detections from different shots, and 0 when no cluster spans two shots.
We report MIR here rather than in \cref{tab:main} because it measures faces only (two of our 25 protagonists are an animal and a robot), it rewards videos that repeat nearly the same frames across shots, and it does not penalize a video that breaks down (Self-Forcing 0.400). Same protagonist is used in the main paper.
With our memory, MIR is higher on both MiniMax-H3 generators (\cref{tab:supp-mir}): by $+0.005$ (RAVEN) and $+0.049$ (LongLive-H3) over the 25 stories, and by $+0.051$ and $+0.070$ on the nine hand-written stories, though none of these differences is significant. Among StoryMem, StoryEngine, and our offline variants, the offline variant on H3 Ref2VA has the highest MIR (0.403 and 0.455, against 0.385 and 0.423 for StoryEngine).
On LongLive-2.0, image memory lowers MIR from 0.334 to 0.191. LongLive-2.0's built-in memory carries the previous shot's K/V and so keeps some faces across adjacent shots, which MIR rewards. ArcFace embeddings of small generated faces are also noisy (cosine 0.1--0.3 between faces of the same shot below 60\,px, 0.6--0.8 above 100\,px), which lowers MIR for every model at these resolutions.

\begin{table}[ht]
\centering
\caption{MIR (StoryEngine protocol, mean over seeds, one seed for SkyReels-V2 and the non-real-time rows). MiniMax-H3: the runs of \cref{tab:main,tab:supp-mainrows} (image memory with the whole $\Gm$). $\Delta$: difference to the same backbone without our memory ($^\ast$: 95\% interval excludes zero).}
\label{tab:supp-mir}
\setlength{\tabcolsep}{3pt}
\resizebox{\columnwidth}{!}{%
\begin{tabular}{l cc cc}
\toprule
& \multicolumn{2}{c}{25 stories} & \multicolumn{2}{c}{9 hand-written} \\
\cmidrule(lr){2-3}\cmidrule(lr){4-5}
Method & MIR & $\Delta$ & MIR & $\Delta$ \\
\midrule
Causal-Forcing (long-video) & 0.466 & -- & -- & -- \\
SkyReels-V2 & 0.471 & -- & -- & -- \\
Self-Forcing & 0.400 & -- & -- & -- \\
CausVid & 0.383 & -- & -- & -- \\
Rolling Forcing & 0.432 & -- & -- & -- \\
LongLive-2.0 (built-in memory) & 0.334 & -- & 0.242 & -- \\
+ image memory, LLM routing (35B) & 0.191 & $-$0.143$^\ast$ & 0.089 & $-$0.153$^\ast$ \\
+ RL planner (image system) & -- & -- & 0.137 & $-$0.105$^\ast$ \\
RAVEN & 0.360 & -- & 0.364 & -- \\
+ image memory (sink) & 0.377 & $+$0.017 & 0.412 & $+$0.049 \\
+ full system & 0.365 & $+$0.005 & 0.414 & $+$0.051 \\
LongLive-H3 & 0.295 & -- & 0.291 & -- \\
+ full system & 0.344 & $+$0.049 & 0.361 & $+$0.070 \\
StoryMem & 0.356 & -- & 0.335 & -- \\
StoryEngine & 0.385 & -- & 0.423 & -- \\
Ours offline (Wan2.2-5B) & 0.374 & -- & 0.390 & -- \\
Ours offline (H3 Ref2VA) & 0.403 & -- & 0.455 & -- \\
\bottomrule
\end{tabular}}
\end{table}

\paragraph{Imaging drift.}
$\Delta$drift is the VBench imaging quality of a story's last minute minus its first (closer to zero is better). LongLive-2.0 with its built-in memory drifts by $-0.089$ and Self-Forcing by $-0.114$, while every variant of our memory keeps LongLive-2.0 within $-0.011$, and the full system reaches $+0.002$. The other systems drift by $-0.003$ (StoryMem) to $-0.064$ (SkyReels-V2), and RAVEN by $-0.009$ without and $-0.003$ with our memory.

\section{Baselines}
\label{sec:supp-baselines}

\paragraph{Real-time baselines.}
\emph{Rolling Forcing} supports a single prompt, so we split generation per shot, re-cache KV at prompt switches as in LongLive~\cite{yang2025longlive}, and wrap RoPE after 1,024 positions (stories reach 1,764 latent frames). It takes 19.3\,s per shot.
The other Wan2.1-T2V-1.3B models use the same protocol: one continuous rollout per story, 42 latent frames per shot (10.3--10.5\,s at 16\,fps), only the current shot prompt, and one seed set before the first shot. Shot 0 of every model is identical to its official inference.
\emph{Self-Forcing} uses the official sampler (4 steps, 3-frame blocks) with the repository's rolling KV cache of 21 frames (its training context) and no sink. Cached frames are recomputed under each new prompt as for Rolling Forcing, and RoPE wraps after 1,024 frames (18.2\,s per shot).
\emph{Causal-Forcing} is run with the authors' long-video checkpoint, which uses the Rolling Forcing inference code (17.4\,s per shot). Its 5-second chunkwise checkpoint, run like Self-Forcing, breaks down within about two shots (87.6\% of its shots have severe artifacts).
\emph{CausVid} uses the repository's long-video script, with rollouts of up to 21 latent frames, each started from 3 overlap frames, and 3 steps. Rollouts never cross a shot boundary, so a new prompt starts the next rollout and re-encodes the overlap under that prompt (13.4\,s per shot).
\emph{SkyReels-V2} uses DF-1.3B-540P at 960$\times$544 with official settings (30 steps, guidance 6, TeaCache 0.3): shot 0 is text-to-video and each later shot is a video extension of 160 frames (10\,s at 16\,fps) conditioned on the last 17 frames, with its own prompt (87.6\,s per shot, one seed).
As it is far from real time, its results are in \cref{tab:supp-skyreels} rather than in \cref{tab:main}: it is below LongLive-2.0 with its built-in memory in state, location, and revisit, slightly above it in identity and same protagonist, and has the lowest imaging quality of all methods.
All times are measured on one H100 (mean over 15 shots after a warm-up, FlashAttention-2, no compilation) and include decoding. The videos themselves were generated on A100s, and the same timing scripts run on an A100 match the generation times within 1--3\%.
Self-Forcing and the chunkwise Causal-Forcing model degrade once the first frames leave the cache, as in their official single-prompt code, so their judge metrics should be read together with the artifact rate.

\begin{table}[ht]
\centering
\caption{SkyReels-V2 on \bench (online protocol, terse prompts, one seed), with two rows of \cref{tab:main} (three seeds) for reference. Artifacts: share of shots with severe artifacts. Time: seconds per shot on one H100.}
\label{tab:supp-skyreels}
\setlength{\tabcolsep}{2.0pt}
\resizebox{\columnwidth}{!}{%
\begin{tabular}{l ccccc cc c}
\toprule
Method & State & Ident. & Same & Loc. & Rev. & Imaging & Artif.\down & Time \\
\midrule
SkyReels-V2 (DF-1.3B, 960$\times$544) & 30.8 & 49.2 & 18.9 & 46.7 & 10.5 & 0.552 & 25.0 & 87.6 \\
LongLive-2.0 (built-in memory) & 38.2 & 44.1 & 18.1 & 68.8 & 15.0 & 0.667 & 7.8 & 15.8 \\
\rowcolor{oursrow} \method (full system) & 64.5 & 89.0 & 46.8 & 98.0 & 55.4 & 0.729 & 6.0 & 16.0 \\
\bottomrule
\end{tabular}}
\end{table}

\begin{table}[ht]
\centering
\caption{Non-real-time systems on \bench (online protocol, terse prompts, 25 stories, one seed) and our real-time full system (three seeds). Shots last 5\,s (StoryMem), 6\,s (StoryEngine, ours offline on Wan2.2-5B), 8\,s (H3 Ref2VA), and 10.5\,s (real time). Time: seconds per shot on one H100 (four for H3 Ref2VA), including API calls. For H3 Ref2VA it is generation only, and the per-shot prompt rewrite adds about 25\,s.}
\label{tab:supp-offline}
\setlength{\tabcolsep}{2.0pt}
\resizebox{\columnwidth}{!}{%
\begin{tabular}{l ccccc cc c}
\toprule
Method & State & Ident. & Same & Loc. & Rev. & Imaging & Artif.\down & Time \\
\midrule
StoryMem (Wan2.2-A14B) & 50.4 & 80.7 & 50.9 & 68.2 & 26.4 & 0.697 & 4.2 & $\sim$880 \\
StoryEngine (Wan2.2-5B) & 69.2 & 96.0 & 67.9 & 96.9 & 70.7 & 0.732 & 3.5 & 432 \\
Ours offline (Wan2.2-5B) & 78.5 & 95.2 & 66.7 & 96.2 & 55.0 & 0.719 & 2.9 & 253 \\
Ours offline (H3 Ref2VA) & 78.0 & 94.6 & 67.0 & 94.6 & 42.7 & 0.728 & 0.5 & 556 \\
\midrule
\rowcolor{oursrow} \method, real time (LongLive-2.0) & 64.5 & 89.0 & 46.8 & 98.0 & 55.4 & 0.729 & 6.0 & 16.0 \\
\bottomrule
\end{tabular}}
\end{table}

\paragraph{Non-real-time systems.}
\emph{StoryMem}~\cite{zhang2025storymem} keeps keyframe memory in Wan2.2-A14B and runs with its official setting (5.06\,s shots, offloading on one H100).
\emph{StoryEngine}~\cite{wang2026storyengine} is an agent system on our Wan2.2-TI2V-5B backbone, run under our protocol: Gemini 3.8 Flash~\cite{google2026gemini} plans, GPT Image~\cite{openai2025gptimage} draws keyframes, and Wan2.2-TI2V-5B generates 6\,s shots. Prompts are revealed one at a time, and its video repair and best-of-$N$ selection are disabled.
\emph{Ours offline} swaps in the API agent (Gemini 3.8 Flash, GPT Image) with synchronous refinement and generates either with H3 Ref2VA (50 steps, up to six references) or with Wan2.2-TI2V-5B image-to-video at StoryEngine's settings (1280$\times$704, 145 frames at 24\,fps, 50 steps, guidance 5, shift 5). For Wan, GPT Image draws a keyframe from the same references and rewritten prompt, and the shot description becomes the text prompt. Unlike StoryEngine, the previous shot's last frame is not used.
These systems spend 28--114$\times$ more time per second of video than our real-time system (\cref{tab:supp-offline}).
On the same Wan2.2-5B backbone and keyframe model, our offline variant beats StoryEngine by 9.3\sig{} state points, ties it in identity, same protagonist, and location, and is 1.7$\times$ faster per shot. StoryEngine leads in revisit ($+15.7$\sig) and slightly in imaging: it anchors every recurring place to a panorama and starts a shot from the previous shot's last frame when they fit, whereas we keep one plate per place and draw every keyframe afresh.
With H3 Ref2VA, the same offline memory reaches about the same state and identity and has the fewest severe artifacts of these systems (0.5\%).
Rolling Forcing, StoryMem, and StoryEngine are released for research or non-commercial use. SkyReels-V2, StoryMem, StoryEngine, and the offline variants use one seed because of cost.

\paragraph{Single-prompt quality (VBench-Long).}
Does memory hurt the base model when nothing needs remembering?
On 42 prompts of VBench (every sixth of the standard list), each generated for about 64\,s as six shots of the same prompt with the image system, against LongLive-2.0's official single-prompt rollout (seed 0, VBench-Long protocol with 2\,s clips), subject, background, and motion consistency are unchanged ($-0.4$, $+0.2$, and $+0.0$ points), aesthetic and imaging quality rise ($+5.0$\sig{} and $+3.5$\sig) because they do not decay over the minute, and dynamic degree falls by 7.5 (not significant).
The small consistency losses come only from clips that straddle one of our 10.5\,s shot boundaries, and LongLive-2.0's own multi-shot mode with the same prompt loses as much dynamic degree ($-6.2$\sig), so they come from cutting into shots rather than from memory.

\section{Robot Videos}
\label{sec:supp-plug}

The agent only decides what the generator sees, so its memory can serve other autoregressive video models, such as robot video models.
All scores are blind judgments of Gemini 3.8 Flash on each arm's own frames, with bootstrap intervals over scenes or episodes.

\paragraph{Close-up recovery (\cref{fig:robot}).}
In robot videos, a camera that stays near an object, or a closed container, hides things that must later come back. Here the memory pictures are frames that the camera has already observed, or crops of them, so no image model is used (\cref{tab:supp-robot}).
The robot video model of Long-WAM~\cite{huang2026longwam} (Robot-M) is LongLive-2.0 trained further on robot and egocentric videos. We run it as released, with a rolling KV cache of about 10.5\,s and no sink.
From each of the seven first frames that ship with it, we synthesize a wrist-camera observation (a 2.5\,s zoom to the task's target object, then a close-up, 13.2\,s in total) and write it into the cache, so the window holds only the close-up.
The model then continues for 5.3\,s under the instruction followed by ``the camera moves back and the whole scene is visible again.''
Our memory is built from the first frame before the close-up: the 35B model of the main text names the target object, returns its box, and writes a card of at most 20 words. The first frame and a crop around the box enter the sink as still blocks, and the card is appended to the prompt.
Without memory, the model turns to a different scene. With our memory, the object is back in 96.4\% of the last frames (60.7\% without), and the same-object score rises by 2.36\sig{} points, in all seven scenes.
Given the same observation, pictures, and card as references, the offline MiniMax-H3 Ref2VA (50 steps) keeps the object even without memory but continues into a different room. Our memory restores the scene (scene consistency from 2.9 to 7.0 of 10, $+4.1$\sig) and raises the same-object score by 1.36\sig{} points.

\begin{table}[t]
\centering
\caption{\textbf{Close-up recovery on robot video models.} \emph{Score} (0--10, blind Gemini 3.8 Flash): the same object in the last frame. \emph{Visible}: the object is in view (\%). $n$: runs over 7 scenes. $\Delta$: ours minus the original model, with a bootstrap 95\% interval.}
\label{tab:supp-robot}
\setlength{\tabcolsep}{2pt}
\resizebox{\columnwidth}{!}{%
\begin{tabular}{l c cc c cc}
\toprule
& & \multicolumn{3}{c}{Score} & \multicolumn{2}{c}{Visible (\%)} \\
\cmidrule(lr){3-5}\cmidrule(lr){6-7}
Generator & $n$ & Orig. & Ours & $\Delta$ [95\% CI] & Orig. & Ours \\
\midrule
Robot-M & 28 & 1.86 & 4.21 & $+$2.36 [1.57, 3.18] & 60.7 & 96.4 \\
MiniMax-H3 Ref2VA & 25 & 5.32 & 6.68 & $+$1.36 [0.83, 1.96] & 100 & 100 \\
\bottomrule
\end{tabular}}
\end{table}

\paragraph{Long-horizon robot tasks.}
Generated robot data needs long videos in which the robot works through many steps while parts of the scene leave the view and come back.
We take the first frames of 30 validation clips of RoVid-X~\cite{deng2026rethinking}, which cover one and two arms, grippers and hands, and head-mounted and fixed cameras.
For each first frame, Gemini 3.8 Flash drafted a script of eight 8\,s instructions (64\,s in total), which we then edited by hand.
The first instructions finish the clip's own task. Each script then hides one or two things and brings each back one to four instructions later. A head camera turns away and back (20 return moments), or a drawer, fridge, box, or lid is closed and reopened, or an object is covered and uncovered (21 moments).
An instruction that brings something back never describes it, for example ``pull the top left drawer open again and reach in for what you put there,'' so only memory can tell the generator what to show.
Each segment continues from the last frame of the previous one, so that the 64\,s video is meant to be one continuous take, although the generator can still cut on its own. Without memory, this frame is all that the generator knows about the scene.
Unlike the main text, where the memory for a shot must be ready before its prompt arrives (C1), each instruction here is available one segment before it plays, as in a scripted robot task, and the agent uses it to choose pictures and rewrite it. This section therefore tests observation-based memory on robot videos, not C1.

\paragraph{Agent.}
Unlike the open-loop agent of the main text, this agent observes the video, because a robot acts in one continuous scene rather than in separate shots. Where an object ends up, or whether a drawer is still open, depends on the generated video rather than on the script, so the memory has to be read from that video. As in the close-up test, it is an observation-based variant: its memory pictures are frames that the robot's camera has already shown, or crops of them, so it needs no image model.
After each segment, the agent looks at three of its frames. It describes the view, lists the visible objects with their position and look, and records whether each container is open or closed and what it holds. It also picks a frame to keep, a frame that shows the inside of an open container, and boxes for up to two objects.
Code turns these observations into two parts of memory. A record of objects and containers keeps an object that was put into a closed container as hidden inside it. A bank of pictures holds one view per segment, container interiors, and the newest crop of each object, each with a caption and the segment it comes from.
Before each segment, a router answers one narrow question about the next instruction: does something come back into view, because the camera turns back, a container is reopened, a cover is lifted, or an object is taken out? It agrees with the scripts' annotations in 188 of 210 segments.
Only when something comes back, the agent sees the bank as thumbnails and lists every picture of the returning place, container, or object. It pins the newest matching view and the newest close view (at most two pictures, \cref{fig:supp-robot-long}) and rewrites the instruction so that it names what comes back and how it looked when last seen.
Otherwise it pins nothing and only rewrites the instruction to name the objects with their look.
Pinning only on returns matters: pictures pinned in almost every segment make the generator restore old states, for example a tube put into a drawer reappears on the cabinet.
The agent was developed on one of the 30 episodes and then kept fixed for all runs.

\begin{table}[t]
\centering
\caption{\textbf{Long-horizon robot videos} (30 episodes, whole 64\,s videos, blind Gemini 3.8 Flash). \emph{Score}: consistency with the last appearance (0--10) at the 41 return moments, with the paired difference to the original model and a bootstrap 95\% interval over episodes. \emph{Visible}: back in view, from the same judgment as the score. \emph{Correct}: in view with identity and position both at least 7, from a separate identity and position judge, scored independently of \emph{Visible}. \emph{Progress}: instruction following (0--10, 210 segments). \emph{Drift}: segment ends with a robot identity score of at most 5. The second seed is compared with the single run of the original model.}
\label{tab:supp-robot-long}
\setlength{\tabcolsep}{2.5pt}
\resizebox{\columnwidth}{!}{%
\begin{tabular}{ll cc ccc c}
\toprule
& & \multicolumn{2}{c}{Score} & Visible & Correct & & Drift \\
Generator & Memory & & $\Delta$ [95\% CI] & (\%) & (\%) & Progress & (\%) \\
\midrule
\multirow{3}{*}{FastH3-Omni} & none & 3.98 & -- & 43.9 & 19.5 & 5.67 & 30.5 \\
& ours & 5.59 & $+$1.61 [0.89, 2.26] & 75.6 & 56.1 & 5.84 & 21.9 \\
& ours, seed 2 & 5.63 & $+$1.66 [0.84, 2.39] & 80.5 & 48.8 & 5.88 & 27.1 \\
\midrule
\multirow{2}{*}{H3 Ref2VA} & none & 3.54 & -- & 29.3 & 4.9 & 5.62 & 39.9 \\
& ours & 5.51 & $+$1.98 [1.40, 2.51] & 65.9 & 43.9 & 5.42 & 28.4 \\
\bottomrule
\end{tabular}}
\end{table}

\begin{figure*}[t]
\centering
\setlength{\tabcolsep}{1pt}
\newcommand{\rl}[1]{\includegraphics[width=0.225\linewidth]{fig/robot_long/#1.jpg}}
\newcommand{\rlab}[1]{\multicolumn{4}{l}{\rule{0pt}{9pt}\footnotesize #1}}
\begin{tabular}{cc@{\hspace{5pt}}cc}
\multicolumn{2}{c}{\small Original model} & \multicolumn{2}{c}{\small Ours} \\
{\footnotesize last seen} & {\footnotesize when it comes back} & {\footnotesize pinned memory picture} & {\footnotesize when it comes back} \\
\rlab{Printer (FastH3-Omni), instruction 8: ``Lift the scanner lid again with the left hand.''} \\
\rl{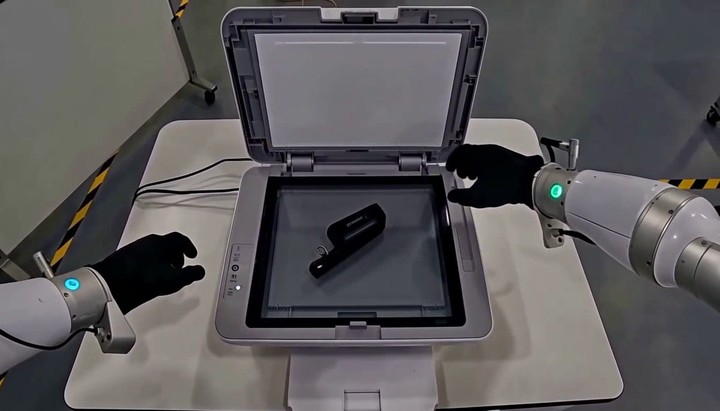} & \rl{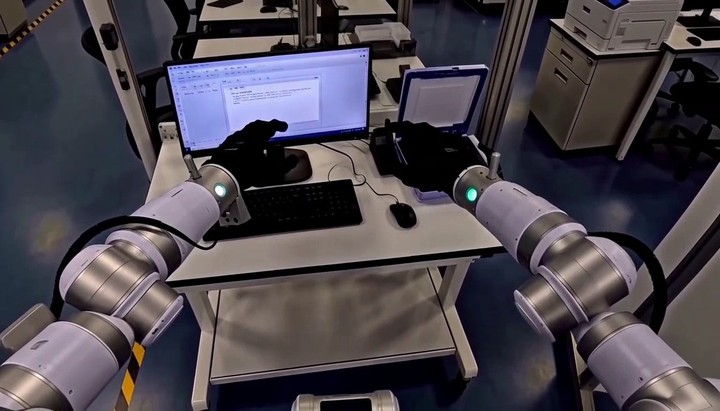} & \rl{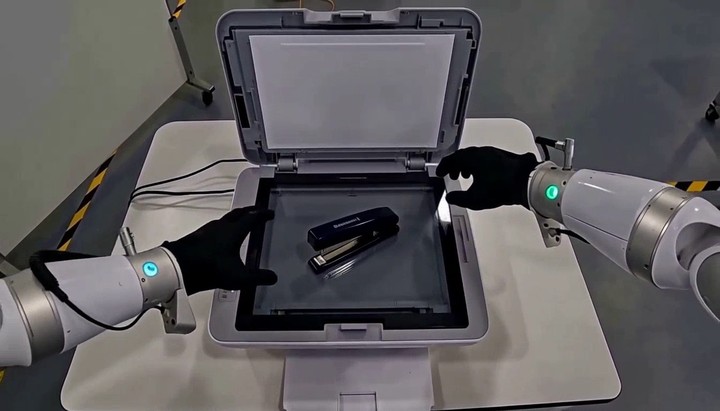} & \rl{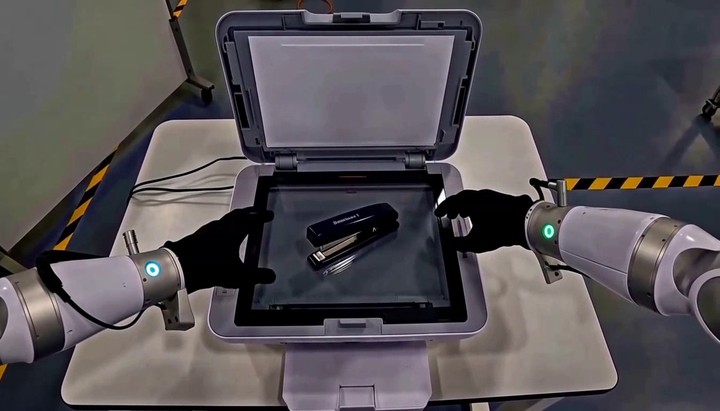} \\
\rlab{Fridge (FastH3-Omni), instruction 7: ``Turn the camera back toward the refrigerator and pull the upper door open again.''} \\
\rl{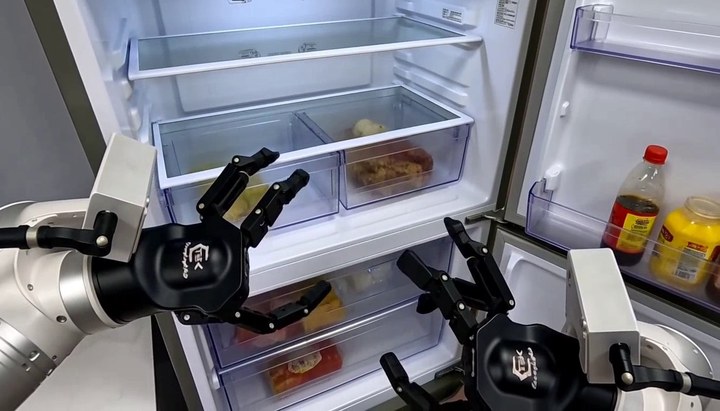} & \rl{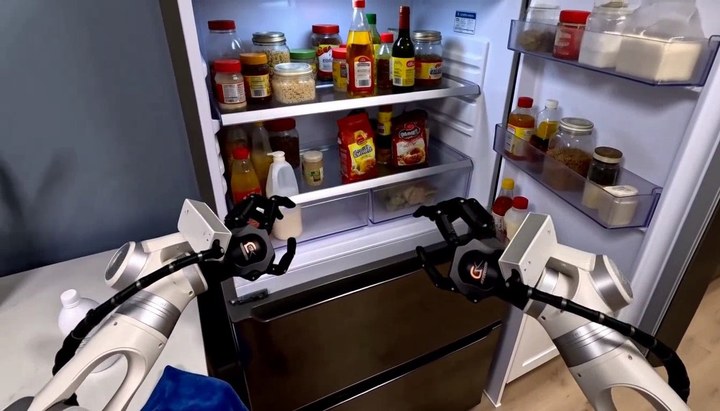} & \rl{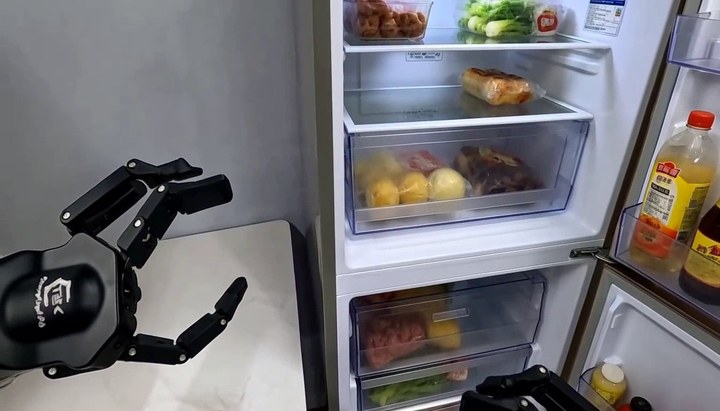} & \rl{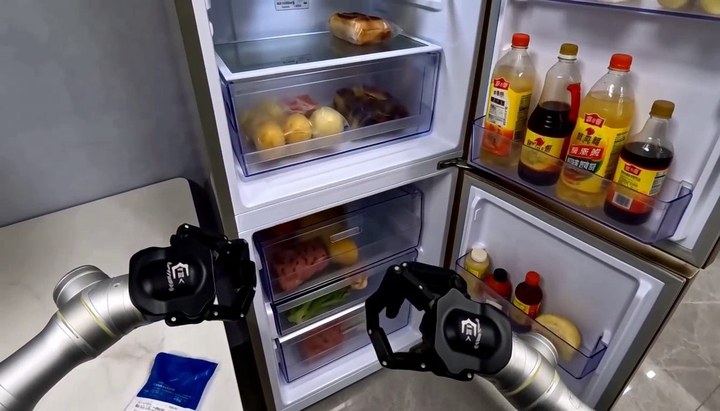} \\
\rlab{Fan (FastH3-Omni, second seed), instruction 6: ``Camera turns back to the center table, re-centering on the electric fan.''} \\
\rl{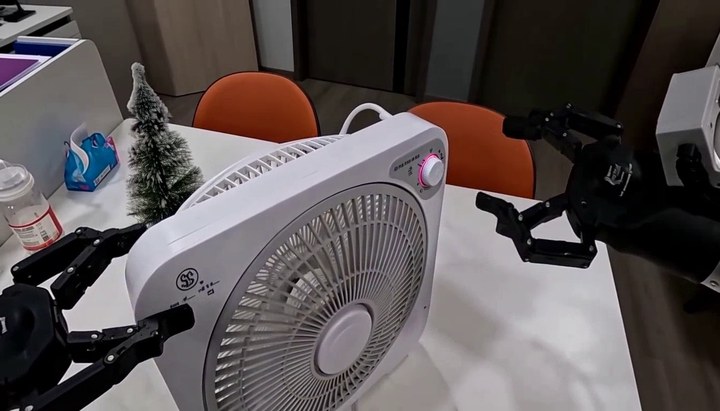} & \rl{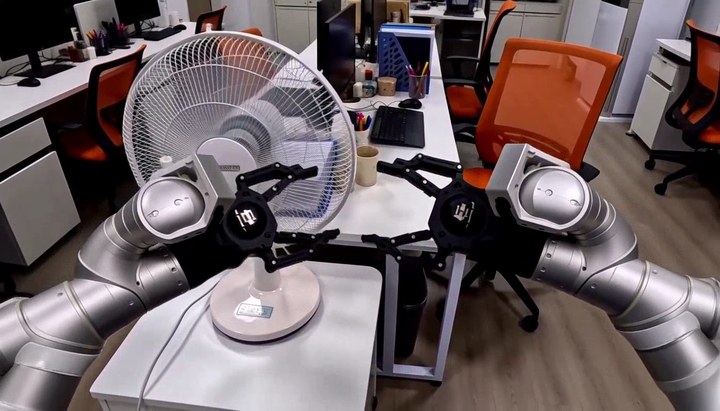} & \rl{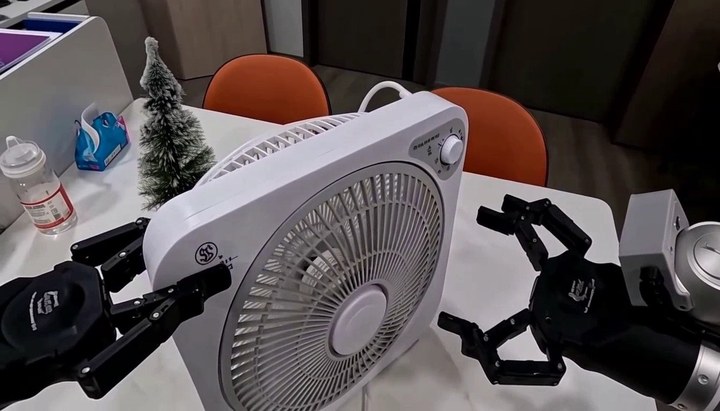} & \rl{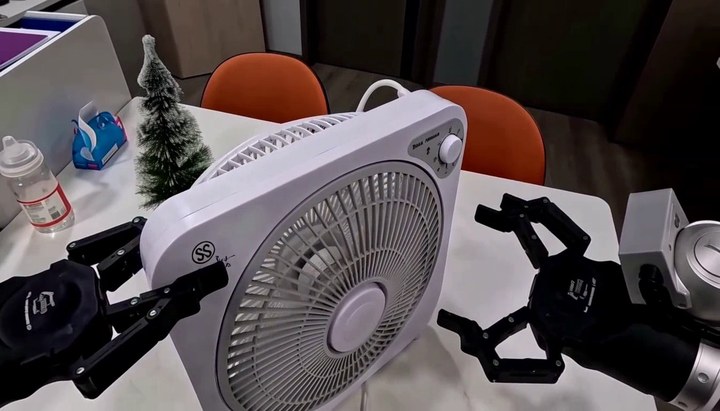} \\
\rlab{T-shirt (H3 Ref2VA), instruction 7: ``Tilt the camera up towards the balcony railing where the white t-shirt was hung.''} \\
\rl{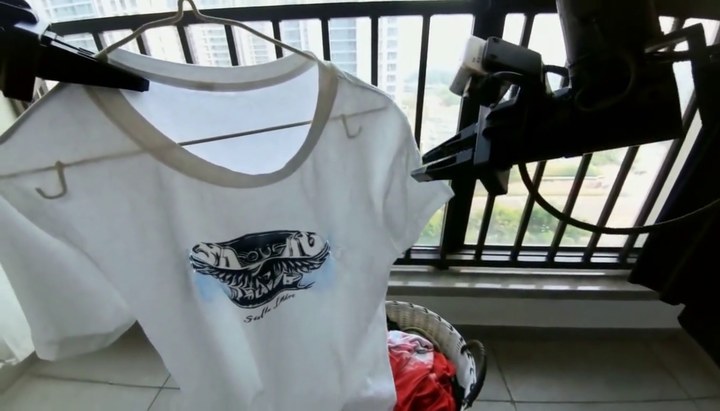} & \rl{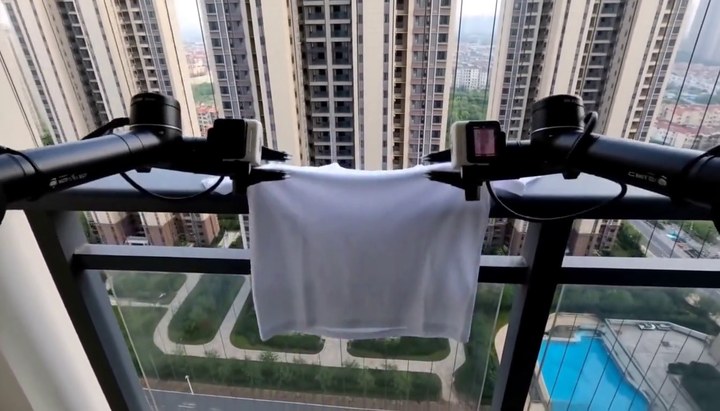} & \rl{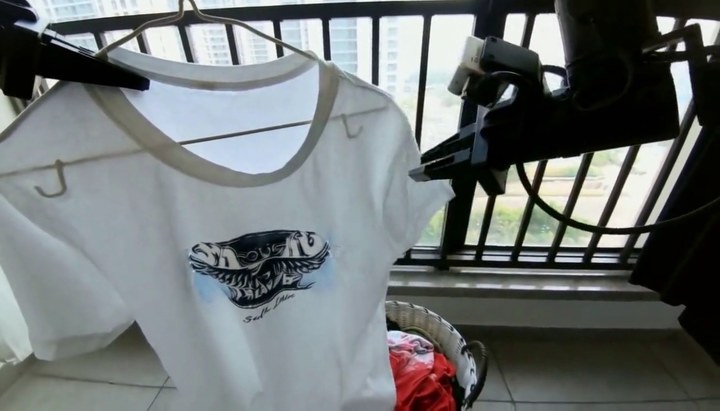} & \rl{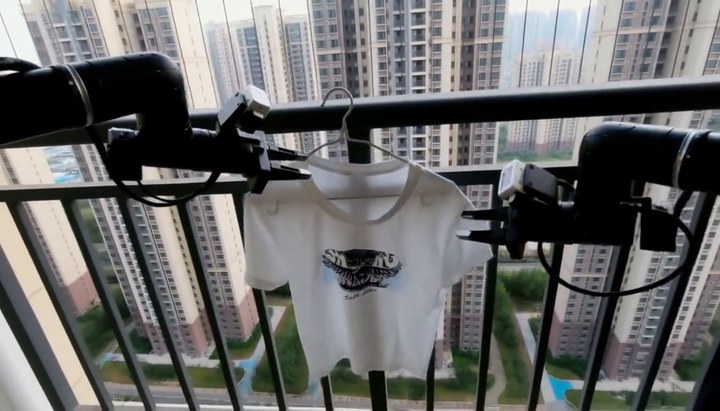} \\
\end{tabular}
\caption{\textbf{Return moments in long-horizon robot videos}, one per row. \emph{Pinned memory picture}: the view our agent pinned for the return, a frame of its own video (object crops not shown). \emph{Last seen}: the original model's frame from the same point of its video. \emph{When it comes back}: the end of the returning segment (fridge: the end of instruction 8).}
\label{fig:supp-robot-long}
\end{figure*}

\paragraph{Generators.}
\Cref{tab:supp-robot-long} uses two generators from earlier sections: FastH3-Omni, as in the real-time runs of \cref{sec:supp-speed}, and H3 Ref2VA, as in the offline variant of \cref{sec:supp-baselines}.
In the real-time system, FastH3-Omni~\cite{fastvideo2026fasth3} (8 steps) generates each segment, and Qwen3.6-35B-A3B~\cite{qwen2026qwen36} runs the agent. The pinned pictures and then the latest frame enter the sink, and the new segment starts from the latest frame. A fixed preamble of the prompt says that the earlier pictures were recorded in the same video, names what each one shows, and states that the target continues from the latest frame without a cut.
While segment $i{-}1$ renders, the agent observes segment $i{-}2$ and prepares segment $i$ from instruction $i$. The quality runs used A100 GPUs.
In the offline system, MiniMax-H3 Ref2VA~\cite{minimax2026h3} (50 steps) generates each segment with the pinned pictures and the latest frame as references, and Gemini 3.8 Flash runs the agent. It observes segment $i{-}1$ before segment $i$ renders and crops objects with Grounding DINO~\cite{liu2024groundingdino} boxes for their names.
The original model receives the latest frame and the user's instruction in the same layout. Both arms share the first segment and use the same seed for every later segment. The second-seed run changes the seeds after the first segment.

\paragraph{Real time and parallelism.}
The closed loop still runs in parallel with generation. The agent observes decoded frames and writes memory while the generator renders on other GPUs, and this memory enters only when the next segment starts, never within a segment. How early the agent can observe depends on the generator. FastH3-Omni is bidirectional and returns a segment only when all of it is decoded, so when segment $i$ starts, the newest frame that the agent has seen is 8\,s old. LongLive-H3 generates a segment chunk by chunk and decodes each chunk as it is generated, so the agent can observe the first half of segment $i{-}1$ while the second half is still being generated, and the newest frame it has seen is 4\,s old.
We timed both loops on six episodes, with the generator on eight B200 GPUs and the agent on two A100 GPUs. With FastH3-Omni, the agent (2.3\,s on average) never makes the generator wait. With the last decoded frame passed straight to the next request, each 8\,s segment takes 7.6\,s on average, and every segment is ready before it is due to play. With LongLive-H3 on five of the GPUs, the first half of a segment is decoded 3.3\,s after the segment starts and the agent needs 1.9\,s, so the generator usually waits about 0.5\,s for it at the boundary. The loop still runs in real time, at 5.8\,s per 8\,s segment on average (at most 7.7\,s), because generation is faster than playback. We time LongLive-H3 but do not score it. It is not trained on robot videos, and with or without memory it does not keep the input robot, whose grippers, for example, turn into humanoid hands early in the video.

\paragraph{Results.}
\Cref{tab:supp-robot-long} compares whole 64\,s videos at the 41 return moments and over all 210 generated segments.
Our memory raises the consistency of what comes back with its last appearance from 3.98 to 5.59 with FastH3-Omni ($+1.61$\sig, a second seed $+1.66$\sig) and from 3.54 to 5.51 with H3 Ref2VA ($+1.98$\sig).
The hidden thing is back in view at 76\% instead of 44\% of the return moments with FastH3-Omni, and back with the same identity and position at 56\% instead of 20\% (H3 Ref2VA: 44\% instead of 5\%).
Camera returns gain the most ($+2.55$\sig{} and $+2.35$\sig), because the pinned view shows the whole place. Reopened containers gain less ($+0.71$, not significant, and $+1.62$\sig): in four of the nine FastH3-Omni reopenings that score below 5, the generator never opens the lid, lifts the cover, or takes the object out, which memory cannot fix.
Task progress stays level (neither difference significant), and segments in which nothing comes back revert to an earlier state about as often as without memory, because nothing is pinned there.
The pinned views also show the robot, so its look stays closer to the input image (same-robot score 7.84 \vs 7.05\sig{} with FastH3-Omni and 7.36 \vs 6.53\sig{} with H3 Ref2VA).
\Cref{fig:supp-robot-long} shows four return moments.

\paragraph{Failure cases.}
The remaining failures fall into three groups.
First, the stored picture can be wrong. The agent sees only three frames per segment, so in one drawer episode it kept the interior from before the tube was put in and pinned that picture at the reopening.
Second, the generator can follow a pinned picture too literally. When the pinned view shows an earlier open state, FastH3-Omni may briefly reopen a lid that was closed, and it sometimes jumps to the pinned frame instead of turning toward it.
Third, the robot itself drifts slowly over the minute, for example from a gripper to a hand. With memory, its identity score is at most 5 in 21.9\% of segment ends (30.5\% without memory).

\section{Entity Evolution and Limitations}
\label{sec:supp-limits}

\begin{figure*}[t]
  \centering
  \includegraphics[width=0.9\linewidth]{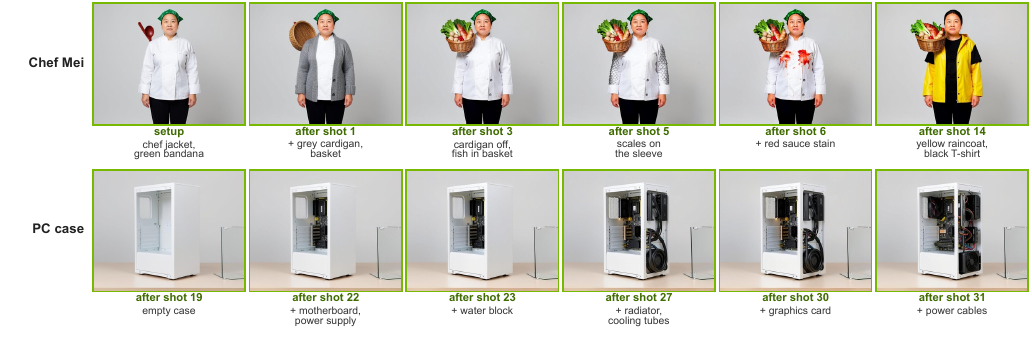}
  \caption{\textbf{Entity evolution} in the RL planner's memory (seed 0): the first image of each version of Chef Mei (``Chef Mei's Longest Day'') and of a PC case (40-shot GPU build, up to shot 31), labelled with the prompt after which it was committed and the state it adds.}
  \label{fig:supp-evolution}
\end{figure*}

\paragraph{Entity evolution.}
\Cref{fig:supp-evolution} shows how the entity images of the RL planner's memory change over a story: a new version is drawn as an edit of the previous image when a prompt states a persistent change, and it shows only persistent attributes on a plain backdrop, with pose and action left to the next prompt.

\paragraph{Failure cases and limitations.}
Memory follows the story up to the previous prompt, because it is built before the next prompt arrives (C3). Two cases follow from this.
A persistent change that the current prompt states for the first time, such as a character shaving, is not yet in the memory images of that shot, which can keep the earlier look until the agent has drawn the change for the following shots.
A recalled place comes with the picture of its last visit. If a character has changed since, the cards state the new look, but an attribute that no card names can follow the earlier picture until the place is drawn again.
Re-injecting updated memory within a shot, once its prompt has arrived, would address both cases and is left to future work.

\emph{Framing.} For every shot, the judge also decides whether the shot size, camera angle, and main subject agree with the planned camera and content (\texttt{framing\_match} in the judge prompt of \cref{sec:supp-bench}). Re-judging agrees on 94.8\% of shots, and the second judge agrees with $\kappa = 0.71$. None of the compared works scores framing.
On LongLive-2.0, framing rises with our full system from 58.5\% to 65.9\% of shots (three seeds), but planned close-ups, a quarter of all shots, fall from 75.5\% to 63.7\%. Memory images show whole entities in medium or wide views and pull the generator toward that shot size, so 78\% of the remaining misses are rendered as medium shots.
Memory is built before the prompt (C3), so it cannot be drawn at the shot size of the coming shot, whereas StoryEngine and our offline variants, which draw a keyframe or references after each prompt, reach 74--95\% framing.
Long text also competes with camera words: cards for every injected member cost 22.4\sig{} framing points alone and 11.9\sig{} with images, whereas cards for the entities that the prompt names cost none ($-0.7$ with the 35B planner, not significant).

\emph{Other limitations.} Memory images can also pull some scene changes toward the previous place. Generators that attend to them in every chunk, such as MiniMax-H3, are affected most: with the whole $\Gm$ they lose moves to new places, which foreground $\Gm$ restores (\cref{tab:supp-fgg}).
On MiniMax-H3, memory K/V also depend on the prompt and cannot be prefilled before it arrives (\cref{sec:speed}).
Real time on LongLive-H3 needs five B200s for generation. On H100 it needs eight, and an H100 agent delivers most updates one shot late (\cref{sec:supp-speed}).
The judge compares shots with text descriptions, which may favor entity cards, but same protagonist, which compares two shots, also rises with cards.

\section{User Study}
\label{sec:supp-userstudy}

\begin{figure}[t]
  \centering
  \includegraphics[width=\linewidth]{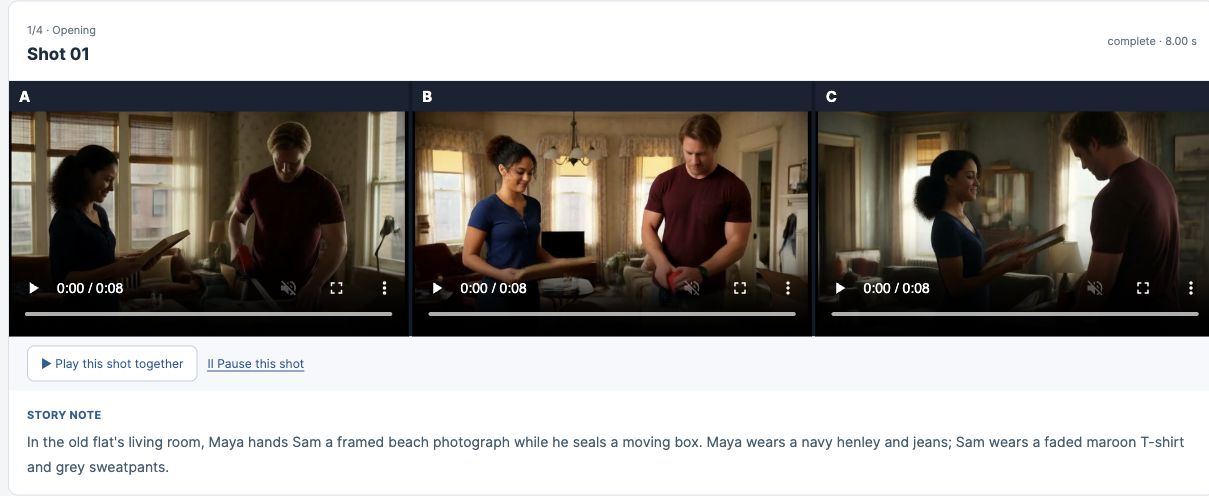}
  \caption{\textbf{Study interface} (LongLive-H3 page, first of four rows): anonymous versions A/B/C with a synchronized play button and a story note. The four criterion questions follow the rows. No participant data are shown.}
  \label{fig:supp-userstudy-ui}
\end{figure}

\paragraph{Setup.}
Thirty-two volunteers from our research community took part, recruited by convenience sampling.
Each judged ten stories of \bench: five comparisons on LongLive-2.0 (its built-in memory \vs our full system) and five on LongLive-H3 (independent shots, K/V carried across shots, and our full system), with the videos of \cref{tab:main} at seed 0.
A comparison shows four shots spread over the story (first, one third, two thirds, and last), with the anonymous versions side by side in a rotating order and a short note on what happens between the shots (\cref{fig:supp-userstudy-ui}). Method names and memory images are not shown.
For each of four criteria, identity consistency, persistent state, location correctness, and visual quality, participants pick the best version, mark a tie, or answer Cannot judge (missing visual evidence or playback failure).

\paragraph{Ethics.}
Participation was voluntary and anonymous and could be stopped at any time. Every response includes informed consent, and the form stored no personal data. No formal IRB review was conducted, as the study was a minimal-risk perceptual evaluation of generated videos.

\paragraph{Results.}
We report the share of best-version votes: a version chosen alone receives one vote, tied versions share it, Cannot judge answers (11 of 1{,}280) are left out, and shares are averaged over the 25 stories. The 95\% intervals are percentiles of 4{,}000 bootstrap draws that resample participants and stories independently.
On LongLive-2.0, against its built-in memory, our full system receives 98.8\% $[95.7, 100]$ of the votes for persistent state, 100\% for identity, 93.8\% for location, and 100\% for visual quality.
On LongLive-H3, in the three-way choice, it receives 97.8\% $[93.2, 100]$ for persistent state, 96.4\% for identity, and 94.0\% for location, while K/V carried across shots receives at most 1.3\% on any criterion.
Visual quality on LongLive-H3 is split between independent shots (55.1\%) and ours (43.6\%), a difference of $-11.4$ points with a 95\% interval of $[-31.3, 9.4]$, and more than half of these answers are ties.
It covers four shots per story and one seed, with a convenience sample and incomplete counterbalancing (21 of 30 layouts), so it complements rather than calibrates the automated metrics.

\end{document}